\documentclass[twoside]{article}

\usepackage[accepted]{aistats2023}
\usepackage{subcaption}
\usepackage{algorithm}
\usepackage{algorithmic}

\usepackage[round]{natbib}

\renewcommand{\cite}{\citep}

\usepackage{hyperref}

\usepackage{amsfonts, amsmath, bm}
\usepackage{optidef}
\usepackage{tikz}

\newcommand{\E}{\mathbb{E}}

\newcommand{\R}{\mathbb{R}}

\def\vd{{\bm{d}}}

\def\vn{{\bm{n}}}

\def\vr{{\bm{r}}}
\def\vs{{\bm{s}}}

\def\vv{{\bm{v}}}

\def\vx{{\bm{x}}}
\def\vy{{\bm{y}}}
\def\vz{{\bm{z}}}
\def\mA{{\bm{A}}}

\def\mI{{\bm{I}}}

\def\gX{{\mathcal{X}}}

\def\gZ{{\mathcal{Z}}}

\DeclareMathOperator*{\argmax}{arg\,max}
\DeclareMathOperator*{\argmin}{arg\,min}

\newcommand{\noise}{\vn}
\newcommand{\Noise}{N}
\newcommand{\enc}{E_{\phi}}
\newcommand{\dec}{D_{\theta}}

\newcommand{\coded}{{k}} 
\newcommand{\inpd}{{m}} 
\newcommand{\npts}{{d}} 
\newcommand{\nptstrain}{\npts_{\text{train}}}

\begin{document}

\twocolumn[

\aistatstitle{Deep Joint Source-Channel Coding with Iterative Source Error Correction}

\aistatsauthor{ Changwoo Lee \And Xiao Hu \And  Hun-Seok Kim }

\aistatsaddress{
University of Michigan 
\And
University of Michigan 
\And
University of Michigan 
} ]


\begin{abstract}
In this paper, we propose an iterative source error correction (ISEC) decoding scheme  for deep-learning-based joint source-channel coding (Deep JSCC). 
Given a noisy codeword received through the channel, we use a Deep JSCC encoder and decoder pair to update the codeword iteratively to find a (modified) maximum a-posteriori (MAP) solution. 
For efficient MAP decoding, we utilize a neural network-based denoiser to approximate the gradient of the log-prior density of the codeword space.
Albeit the non-convexity of the optimization problem, our proposed scheme improves various distortion and perceptual quality metrics from the conventional one-shot (non-iterative) Deep JSCC decoding baseline. 
Furthermore, the proposed scheme produces more reliable source reconstruction results compared to the baseline when the channel noise characteristics do not match the ones used during training.
\end{abstract}

\section{INTRODUCTION}

Joint source-channel coding (JSCC) is a problem to find an encoder-decoder pair to transmit a compressible source reliably through a noisy channel.
Deep-learning-based methods have brought advancement for (Deep) JSCC of images  \citep{bourtsoulatze2019deep,yang2021deep,yang2022deep, grover2019uncertainty, choi2019neural} by \textit{learning} a pair of encoder and decoder network models from data.
Deep JSCC encoder and decoder pairs are parameterized by deep neural networks in the form of autoencoders \cite{kramer1991nonlinear, kingma2013auto}. 
The latent variables at the output of the encoder is the transmitted codeword. The receiver uses a paired decoder to reconstruct the original source by decoding the noisy received codeword (i.e., the latent variables corrupted by the channel noise).
The neural network encoder directly maps the source input to a codeword  whose dimension is usually smaller than the input (i.e., signal compression), and the decoder network is trained to approximate the inverse mapping from the noisy codeword to the source.

In this paper, we consider an \textit{asymmetric} communication case where the transmitter (e.g., an Internet-of-Things device) is \textit{power-constrained} while the receiver (e.g., a powerful base station) enjoys \textit{abundant resources}.
The conventional Deep JSCC framework decodes a noisy codeword using the decoder network only once. 
This `one-shot' decoding is efficient in terms of power and time complexity, but it does not provide optimal performance in practice. 
For example, (one-shot) Deep JSCC decoders often fail to produce better quality than conventional (not neural network-based) source and channel coding schemes when the noise power and compression ratio are low \cite{bourtsoulatze2019deep}.
Moreover, this suboptimality inevitably worsens when the channel noise characteristics shift from those of the noise model used for training.

To address these limitations of conventional Deep JSCC, we introduce an \textit{iterative source error correction} (ISEC) scheme that iteratively decodes the received noisy codeword to gradually improve the reconstructed source quality.
ISEC iteratively updates the codeword via gradient ascent by combining the \textit{likelihood} of the current estimate based on the noise statistics and the \textit{prior probability} of the codeword. 
We optimize/update the estimate from the noisy codeword to find the (local) minimum of a target loss using a Deep JSCC decoder.
Accordingly, our proposed method approximately estimates a modified \textit{maximum a-posteriori} (MAP) solution within the set of points around the observed codeword.

One can consider the Deep JSCC decoding as a non-linear inverse problem since the receiver only observes a non-linearly compressed, noisy version of the source.
Thus obtaining an optimal solution requires a proper \textit{prior} structure on the source domain for minimizing the negative likelihood of the source estimate based  on the observed noisy codeword.
To be specific, we use two types of the prior structure.
We first adopt a Deep JSCC decoder network as an image prior, and modify the received noisy codeword by observing the reconstructed (and re-encoded) source at the receiver to correct errors.
Secondly, we impose an \textit{implicit prior} to the codeword space because feeding an arbitrary input to a Deep JSCC decoder may produce meaningless image pixels due to the absence of a prior regularization during the training of Deep JSCC models \cite{grover2019uncertainty}.
Instead of explicitly modeling a prior density function which is difficult to obtain, we formulate ISEC to use the gradient of the log density approximated by a neural network \textit{denoiser} inspired by \cite{kadkhodaie2020solving}.

We demonstrate that the proposed ISEC improves various distortion and perceptual quality metrics outperforming one-shot Deep JSCC decoding of low and high-resolution images in CIFAR-10 and Kodak datasets.
The gain becomes more significant when the characteristics of the channel noise for the evaluation are different from those of the training environment.
The contributions of this paper are as follows:
\begin{itemize}
    \item We propose a new iterative source error correction (ISEC) decoding algorithm for modified MAP JSCC decoding of low- and high-resolution images.
    \item We formulate the ISEC MAP scheme to utilize the gradient of the prior distribution over the \textit{codeword space} modeled by a bias-free denoiser network. 
    \item We experimentally show that  ISEC decoding outperforms the one-shot counterpart in various metrics, especially when the noise statistics  mismatch those used during the training of the Deep JSCC encoder and decoder.
    We also analyze and quantify the impact of different terms in ISEC under various situations.\footnote{ISEC code is available at \url{https://github.com/changwoolee/isec-deep-jscc}}
\end{itemize}

\section{BACKGROUND AND RELATED WORKS}

\subsection{Separate and Joint Source-Channel Coding}
For a source with redundant information, 
a conventional separate source-channel coding (SSCC) approach transmits the source over a noisy channel using a separate source-dependent coding scheme (e.g., JPEG \cite{jpeg} or better portable graphics (BPG) \cite{bellad2018bpg}) followed by a source-independent channel coding scheme (e.g., low-density parity-check (LDPC) code \cite{gallager1962low}).
On the other hand, joint source-channel coding (JSCC) uses a single joint code for source encoding and channel coding.

\subsection{Deep Joint Source-Channel Coding}

Deep joint source-channel coding of images (Deep JSCC of images or shortly Deep JSCC) \cite{bourtsoulatze2019deep, grover2019uncertainty} is a deep-learning-based approach of JSCC to transmit an image $\vx\in\gX$ ($\subset\R^{\inpd}$) over a noisy channel and reconstruct the source at the receiver to minimize the error.  
Unlike SSCC schemes, there is no explicit boundary between source and channel (de)coding in Deep JSCC encoders (decoders).

Deep-JSCC training finds the parameters $\phi$ and $\theta$ of an encoder $\enc:\R^\inpd \to \R^\coded$ and a decoder $\dec:\R^\coded \to \R^\inpd$ modeled by deep neural networks. The received codeword $\vy$ and the reconstructed source at the decoder output $\hat{\vx}$ can be expressed by
 \begin{align*}
    \hat{\vx} = \dec(\vy), \quad	\vy=\enc(\vx)+\noise,  
 \end{align*}
where $\noise$ is the additive  channel noise that follows the distribution $P_\sigma(\Noise)$ parameterized by $\sigma$.
We assume $\coded < \inpd$ (i.e., the encoder compresses the source to a smaller dimension), and without loss of generality, we further assume that the squared norm of codewords is bounded such that for all $\vx \in \gX$,
    $\|\enc(\vx)\|_2^2 \le \coded.$
The signal-to-noise ratio (SNR) in dB is defined by
\begin{align*}
    SNR=10\log_{10}\frac{\|\enc(\vx)\|_2^2}{\|\noise\|_2^2}.
\end{align*}
We use the notation $SNR_{\text{train}}$ and $SNR_{\text{test}}$ to denote the SNR during the training and the test, respectively.
\textit{Channel Per Pixel} (CPP) indicates the inverted compression ratio
\begin{align*}
    CPP = 0.5 \coded / \inpd,
\end{align*}
where the factor of $0.5$ is because of the complex-valued transmission of the codeword that allows transmitting \textit{two} real-valued elements per channel-use as explained in the Deep JSCC literature \cite{bourtsoulatze2019deep, yang2021deep, yang2022deep, kurka2020deepjscc}.

The pair $(\enc, \dec)$ can be optimized/trained jointly with independent and identically distributed (i.i.d.) data points $\vx_1,\ldots,\vx_{\nptstrain}\in\gX$ and i.i.d. channel noise  $\noise_1,\ldots,\noise_{\nptstrain}\sim P_{\sigma_{\text{train}}}(\Noise)$:
\begin{align}\label{eq:deep_jscc_training}
    (\phi, \theta) &:= \argmin_{\phi, \theta} \frac{1}{\nptstrain}\sum_{i=1}^{\nptstrain}\|\vx_i - \dec(\enc(\vx_i)+\noise_i)\|_2^2.
\end{align}

Due to the generalization ability of neural networks, Deep JSCC decoders can often provide a more reliable reconstruction of the image from the noisy codeword than separate source-channel coding (SSCC) schemes when the decoder faces unexpected/mismatched noise characteristics.
It is also shown that under the additive white Gaussian noise with known noise variance, Deep JSCC outperforms SSCC schemes when SNR is low and the source compression ratio is high, whereas the gain diminishes in the opposite scenario (SNR is high and the compression ratio is low).
Deep JSCC framework has been extended to binary symmetric and erasure channels \cite{choi2019neural}, an SNR-adaptive framework \cite{ding2021snr}, rate-adaptive framework \cite{yang2022deep}, OFDM-guided framework \cite{yang2021deep}, and feedback channels \cite{kurka2020deepjscc}.
Also, \citet{grover2019uncertainty} discussed the Deep JSCC framework in the compressed sensing literature, focusing on reducing the number of measurements corrupted by mild noise.
All prior Deep JSCC frameworks (except the schemes designed with feedback channels \cite{kurka2020deepjscc}) employ a \textit{one-shot decoding method} without iteration.
Those one-shot Deep JSCC schemes can benefit from our proposed ISEC framework introduced in Section \ref{sec:proposed_method} to  improve the quality of the reconstructed source by iterations at the receiver without modifying the transmitter or requesting retransmission. Since we assume that abundant power and computational resources (e.g., GPUs) are available at the receiver, improving performance is our primary objective, whereas the increased decoding complexity is a secondary concern in our framework.


\subsection{Compressed Sensing using Generative Models}

For compressed-sensing-based source compression and reconstruction \cite{donoho2006compressed,candes2006near}, deep generative models (DGMs) have emerged as a new type of prior for natural images.
Compressed sensing using generative models (CSGM) employs generative adversarial networks (GANs) \cite{goodfellow2014generative} or variational autoencoders (VAEs) \cite{kingma2013auto}, as first proposed by \citet{bora2017compressed}. 
The intuition behind the approach is to recover the source $\vx$ from a \textit{linear} measurement $\vy=\mA\vx+\noise, \mA\in\R^{\coded\times\inpd}$,  by searching over the range of a (possibly non-linear) generative model $G:\R^r\to\R^\inpd$ which maps a simple distribution $P_{\gZ}$ (e.g., isotropic Gaussian distribution) to the distribution of the sources $P_{\mathcal{X}}$. 
CSGM reconstructs $\hat{\vx}_{\text{CSGM}}$ as follows: 
\begin{align*}
    \hat{\vz}_{\text{CSGM}}&:=\argmin_{\vz\in\gZ} \|\vy-\mA G(\vz)\|^2,\\
    \hat{\vx}_{\text{CSGM}}&:= G(\hat{\vz}_{\text{CSGM}}),
\end{align*}
where $\vz$ is initialized randomly or to zero, then updated by a standard gradient descent method such as Adam \cite{kingma2014adam}.
Although the non-convexity of the optimization problem, CSGMs outperform  conventional sparsity-based compressed sensing solvers on relatively simple source distributions such as human face images \cite{liu2015faceattributes}.
Theoretical analysis and improvement have been reported by \citet{joshi2021plugin} whereas the techniques of inverting the generative models have been investigated in \cite{creswell2018inverting, lei2019inverting,asim2020invertible,jalal2021instance}.

Training DGMs on high-resolution natural images is a challenging task. Therefore, the direct application of CSGM to the Deep JSCC problem for natural images has been limited.
Also, because the optimization process starts from random or zero-valued vectors as the GAN/DGM input, it requires thousands of iterations to converge to one of local optima\cite{bora2017compressed}.
Although CSGM shares some similarities with our proposed ISEC method, there are a few main differences. 1) ISEC recovers the source from \textit{non-linear} measurements $\vy=\enc(\vx)+\noise$ to operate in lower SNR scenarios. 2) ISEC uses a Deep JSCC encoder and decoder as the DGM, trained with a highly complex latent distribution. 3) ISEC initializes the variable to $\vy$, which is already close to the local minima, rather than a random or zero vector as in CSGM. 4) For a modified MAP solution, ISEC utilizes the approximate gradient of the \textit{log prior density} of the codeword obtained by a denoiser neural network.

\subsection{Gradient Estimation of Log Prior using Bias-free Denoiser}

Our proposed decoding algorithm is built on the seminal work by \citet{miyasawa1961empirical}.
Consider an observed point $\vr$ of the original signal $\vs$ corrupted by Gaussian noise with variance $\sigma^2$. Then, the gradient of the log density function $\nabla \log p(\vr)$ can be expressed with the minimum mean squared error (MMSE) estimator $\hat{\vs}(\vr)=\E[\vs|\vr]=\int \vs p(\vs|\vr)d \vs$, satisfying
\begin{align}\label{eq:mmse_logdensity}
    \nabla \log p(\vr) = \frac{\hat{\vs}(\vr)-\vr}{\sigma^2}.
\end{align}
\citet{kadkhodaie2020solving} and \citet{kawar2021snips} used a deep-neural-network-based MMSE estimator, where a DNN is trained to predict the noise instance $\noise$ from the corrupted observation $\vr = \vs + \noise$. 
A common choice for the DNN is a \textit{bias-free} convolutional neural network due to its robustness to the unseen noise variance \cite{mohan2020robust}.
Our ISEC algorithm adopts a similar approach to formulate and solve a modified MAP optimization problem using a denoiser network.
However, unlike prior works, we obtain the gradient of the log prior density in the \textit{codeword} space.


\section{PROPOSED METHOD}\label{sec:proposed_method}


We introduce an iterative decoding algorithm, namely \textit{iterative source error correction} (ISEC), to refine the initial / one-shot output of the Deep JSCC decoder $\dec(\vy)$. 
Given a Deep JSCC \cite{bourtsoulatze2019deep, grover2019uncertainty} encoder $\enc$  and decoder $\dec$ pair, ISEC finds a modified MAP solution in the codeword space via an iterative process.

\subsection{Assumptions}
We suppose both the transmitter and receiver share the same $(\enc, \dec)$ pair, and the receiver observes the $\coded$-dimensional noisy codeword $\vy=\enc(\vx)+\noise$ of the $\inpd$-dimensional source (image) $\vx$ transmitted through the noisy channel. 
Also, the trained decoder network $\dec$ is assumed to be deterministic and injective.
We assume the channel noise follows isotropic Gaussian distribution $\noise \sim p_\sigma(\Noise)=\mathcal{N}(\boldsymbol{0}, \sigma^2 \mI)$.
We further assume that the receiver can estimate the true noise variance $\sigma^2$ after observing the noisy codeword so that the noise variance is available for our decoding algorithm (Algorithm \ref{alg:ISEC_basic}).
The transmitter may or may not have the true noise variance information prior to transmission. 
Moreover, we consider an \textit{asymmetric} communication case where the transmitter is power- and resource-constrained while the receiver has abundant computational resources with an unlimited power constraint for iterative decoding.


\subsection{Maximum A-Posteriori Inference for Deep JSCC}

Given $\vy$, our goal is to find a \textit{maximum a-posteriori} (MAP) solution $\hat{\vz}_{\text{MAP}}$ in the codeword (latent) space $\gZ\subseteq \R^\coded$:
\begin{align}
    \hat{\vz}_{\text{MAP}} &:= \argmax_{\vz\in\gZ} \log p(\vz|\vy), \label{eq:MAP_Opt}
\end{align}
where a reconstructed source $\hat{\vx}_{\text{MAP}}$ given $\hat{\vz}_{\text{MAP}}$ is obtained by $\hat{\vx}_{\text{MAP}} := \dec(\hat{\vz}_{\text{MAP}})$.
That is, we find the reconstruction in the range of the decoder network, which is our first \textit{prior} structure in the \textit{source} domain.
Iterating the time index $t=0,\ldots,T-1$, we update $\vz_t$ by gradient ascent defined as
\begin{align*}
    \vz_{t+1} := \vz_{t} + \eta_t \cdot \nabla_{\vz_t} \log p(\vz_t|\vy),
\end{align*}
where $\eta_t>0$ is a step size and $\vz_0:=\vy$.
Unfortunately, the direct estimation of the gradient of the log posterior density is not available in general.
Thus we consider the following form instead:
\begin{align}
    \nabla_{\vz} \log p(\vz|\vy) = \nabla_\vz \log p(\vy|\vz) + \nabla_\vz \log p(\vz) \label{eq:gradient_bayes}.
\end{align}
Let $\mathcal{R}(\dec)\subset \gX$ be the range of $\dec$.
Since $\dec$ is deterministic and injective, one could write $p(\hat{\vx}|\vz)=\delta(\hat{\vx}-\dec(\vz))$ where $\delta(\cdot)$ is a Dirac delta function.
Consider a marginalized conditional distribution $p(\vy|\vz)$ over $\hat{\vx}\in\mathcal{R}(\dec)$:
\begin{align}
    p(\vy|\vz) &= \int_{\mathcal{R}(\dec)} p(\vy, \hat{\vx} | \vz) \mathrm{d}\hat{\vx} \nonumber\\
   &= \int_{\mathcal{R}(\dec)} p(\vy| \hat{\vx} , \vz) p(\hat{\vx}|\vz) \mathrm{d}\hat{\vx}  \nonumber\\
   &= \int_{\mathcal{R}(\dec)} p(\vy| \hat{\vx} , \vz) \delta(\hat{\vx}-\dec(\vz)) \mathrm{d}\hat{\vx}\nonumber  \\
   &= p(\vy|\dec(\vz)) \nonumber\\
   &= p_\sigma(\vy-\enc(\dec(\vz))).
\label{eq:marginal_dist}
\end{align}
The first term on the right-hand side of \eqref{eq:gradient_bayes} is a likelihood of noise distribution. Using \eqref{eq:marginal_dist}, it can be written in the following form:
\begin{align}
    \nabla_\vz\log p(\vy|\vz) =-\nabla_\vz \frac{1}{2\sigma^2}\|\vy - \enc(\dec(\vz))\|_2^2. \label{eq:ISEC_Likelihood}
\end{align}
As we seek a MAP solution in the domain of the decoder $\dec$, optimizing $\vz$ using the likelihood term is equivalent to finding an optimal $\vx$ in the range of $\dec$. This approach shares a similarity with the CSGM framework \cite{bora2017compressed}.

Since $\vy$ is corrupted by noise, the direction in \eqref{eq:ISEC_Likelihood} does not always lead to reducing the source error $\|\vx - \dec(\vz)\|_2^2$. 
Moreover, no distributional or structural prior is given for the \textit{codeword} (latent) space between the Deep JSCC encoder and the decoder.
Unlike the latent space of DGMs in the CSGM scheme which can be described by a compact set such as a closed ball \cite{bora2017compressed}, the latent space for Deep JSCC networks has a complex structure.
Therefore, instead of finding a solution starting from a random or zero-valued vector, we start from $\vz_0:=\vy$, using the decoder $\dec$  trained to reconstruct $\vx$ which is one of the local optima.
Then, through the iterative inference (test) stage, we further refine the codeword. 
In the proposed scheme, a proper choice of the regularizer is critical  to keep $\vz_t$ in the local region around the noiseless codeword $\enc(\vx)$, especially when the codeword $\vy$ is corrupted by noise with lower SNR than trained.

\begin{figure}[]
    \centering
    \includegraphics[width=\linewidth]{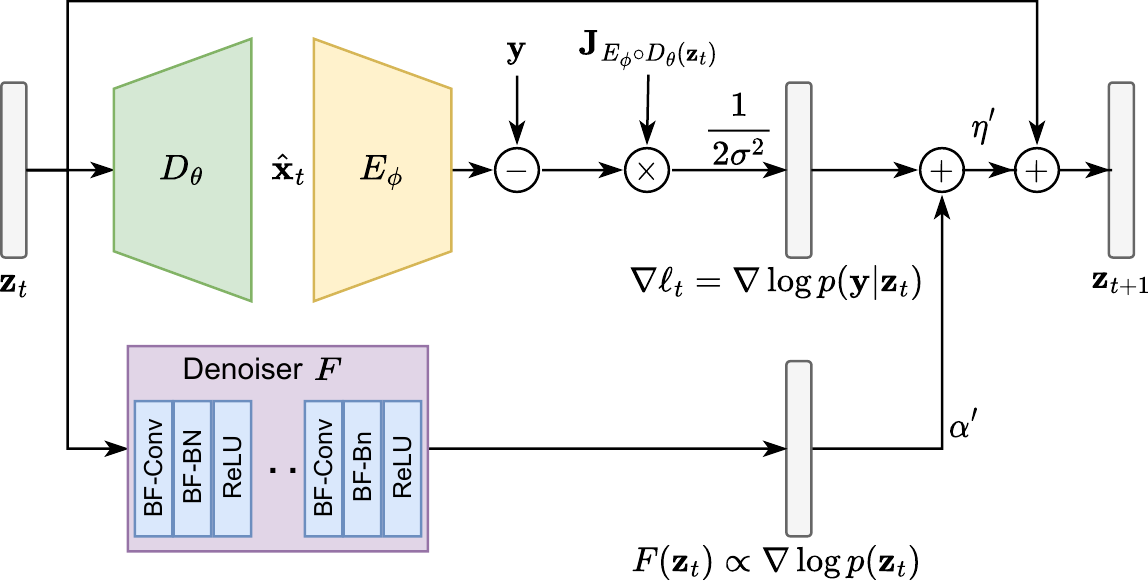}
    \caption{Illustration of ISEC at step $t$ in \eqref{eq:GD_Final}. $\mathbf{J}_{\enc\circ\dec(\vz_t)}$ refers to the Jacobian matrix of $\enc\circ\dec(\cdot)=\enc(\dec(\cdot))$ at $\vz_t$.}
    \label{fig:ISEC_block_diagram}
\end{figure}

\subsection{Denoiser-based Regularizer for Modified MAP Inference}

While a proper regularizer plays a critical role, the absence of a prior structure on the codeword space makes it difficult to find. 
This motivates us to consider an \textit{implicit} prior over the codeword space.
That is, instead of modeling the marginal distribution of the codeword space $p(\vz)$, we focus on estimating the \textit{gradient} of the log density function $\nabla \log p(\vz)$. 
Given $\vz_t$ at the $t$-th gradient ascent step, we know from \eqref{eq:mmse_logdensity} that the MMSE estimation $\hat{\vz}(\vz_t)=\mathbb{E}[\vz | \vz_t]$  can be used to approximate the gradient of the log prior density of $\vz_t$ by
\begin{align}\label{eq:gradient_logprior}
    \nabla \log p(\vz_t) = \frac{\hat{\vz}(\vz_t)-\vz_t}{\sigma_t^2} \approx \frac{F(\vz_t)}{\sigma_t^2},
\end{align}
where $F:\gZ\to\gZ$ is a \textit{denoiser network} operates in the \textit{codeword space}, and $\sigma_t^2$ is the variance of the current estimation $\hat{\vz}(\vz_t)-\vz_t$.
We train the denoiser network to predict the noise instance $\noise$ from the corrupted codeword $\vy=\enc(\vx)+\noise$. See \ref{sec:append_training} for details on training.

To stabilize the optimization process, we propose a \textit{modified} version of the MAP estimator based on the following expression with a parameter set $\{\alpha_t\}_{t=0}^{T-1}$:
\begin{align*}
    \nabla_{\vz_t}\log \tilde{p}_{\alpha_t}(\vz_t |\vy) &:= \nabla_{\vz_t} \log p(\vy|\vz_t) + \alpha_t \nabla_{\vz_t}\log p(\vz_t) 
\end{align*}
so that
\begin{align*}
    \vz_{t+1} &:= \vz_{t} + \eta_t \cdot ( \nabla_{\vz_t} \log p(\vy|\vz_t) + \alpha_t \nabla_{\vz_t}\log p(\vz_t) ).
\end{align*}
By letting $\eta_t = \eta'$ and $\alpha_t=\alpha' \cdot \sigma_t^2$, we arrive at the expression for the $t$-th gradient ascent step:
\begin{align}
    \vz_{t+1} &:= \vz_t + \eta' \cdot \left(-\nabla \frac{1}{2\sigma^2}\|\vy - \enc(\dec(\vz_t))\|_2^2 + \alpha' F(\vz_t)\right)
    \label{eq:GD_Final}
\end{align}
as illustrated in Figure \ref{fig:ISEC_block_diagram}.

\textbf{Choice of hyperparameters} 
Our preliminary experiments show that making hyperparameters adaptive to the SNR \textit{mismatch} between the testing and the trained environment significantly improves the quality of reconstructions.
Based on the trained noise variance $\sigma_{\text{train}}^2$, we adaptively scale the step size $\eta'$ and the regularization parameter $\alpha'$ for the current noise variance $\sigma^2$.
Given $\alpha>0, \eta>0$, and $\delta\ge 0$ which are the same for each model $(\phi,\theta)$ trained for the same codeword dimension and SNR, we adjust the step size $\eta$ and the contribution of the prior $\alpha'$ by
\begin{align}
    \alpha' := \alpha \cdot h(\sigma, \sigma_{\text{train}}, 2), \quad
    \eta':= \eta/ h(\sigma, \sigma_{\text{train}}, \delta),\label{eq:alpha_eta}
\end{align}
where 
\begin{align}
h(\sigma, \sigma_{\text{train}}, \delta) &:= \max\left(0.1,  \left(\frac{\sigma^2}{\sigma_{\text{train}}^2}\right)^{\delta'}\right), \\ 
    \delta' &:= \begin{cases}
    \delta & \text{if }\sigma^2 < \sigma_{\text{train}}^2  \\
    1 & \text{otherwise}
    \end{cases}. \label{eq:delta}
\end{align}

For an extreme case where $\sigma^2 \to \infty$ and $\delta > 0$, $h(\sigma, \sigma_{\text{train}}, \delta) \to \infty$ thus $\alpha' \to \infty$ and $\eta' \to 0$.
That is, the contribution of the likelihood to the gradient ascent becomes less significant and the step size decreases when the current noise variance ($\sigma^2$) is higher than the trained noise variance ($\sigma_{\text{train}}^2$).
Algorithm \ref{alg:ISEC_basic} summarizes the overall procedure of the proposed ISEC MAP algorithm.

\begin{algorithm}[ht]
    \caption{Iterative Source Error Correction}\label{alg:ISEC_basic}
    \textbf{Input}: $\vy$, $\sigma>0$, $\sigma_{\text{train}}>0$ $\eta>0$, $T>0$, $\alpha > 0$, $\delta\ge0$, $\enc, \dec,F$. \\
    \textbf{Output}: $\hat{\vx}$ 
    \begin{algorithmic}[1]
        \STATE $\alpha'\gets \alpha \cdot h(\sigma, \sigma_{\text{train}}, 2)$
        \STATE $\eta'\gets \eta / h(\sigma, \sigma_{\text{train}}, \delta)$ 
        \STATE $\vz_0 \gets \vy$
        \FOR{$t=0,\ldots,T-1$}
            \STATE $\ell_t \gets -\frac{1}{2\sigma^2}\|\vy - \enc(\dec(\vz_{t}))\|_2^2$
            \STATE $\vd_t \gets F(\vz_t)$
            \STATE $\vz_{t+1} \gets \vz_t + \eta' \cdot \left(\nabla \ell_t + \alpha' \vd_t\right)$
        \ENDFOR
        \STATE $\hat{\vx} \gets \dec(\vz_{T})$
        \STATE \textbf{return} $\hat{\vx}, \vz_T$
    \end{algorithmic}
\end{algorithm}

\section{EXPERIMENTS}\label{sec:experiments}

\begin{figure*}
    \centering
    \begin{subfigure}{\linewidth}
    \includegraphics[width=1.0\linewidth]{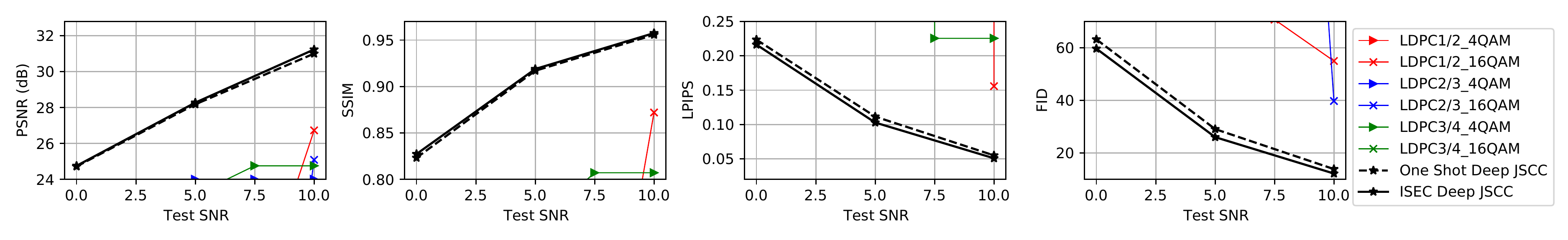}
    \caption{CIFAR-10}
    \label{fig:snr_match_cifar}
    \end{subfigure}
    \begin{subfigure}{\linewidth}
    \includegraphics[width=1.0\linewidth]{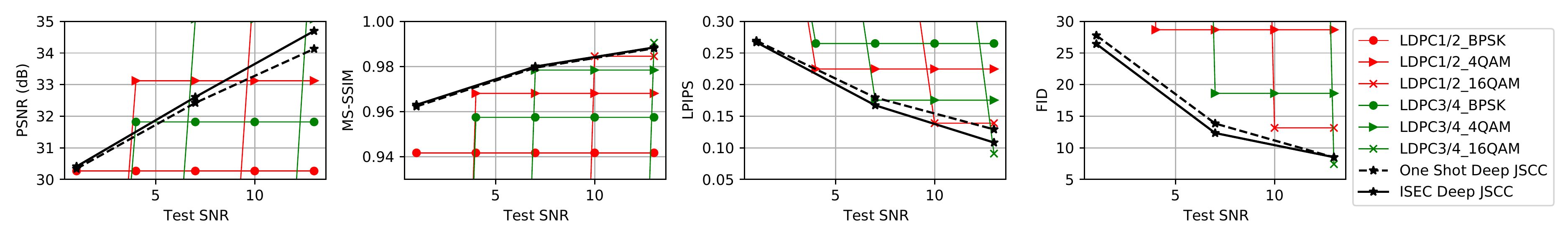}
    \caption{Kodak dataset}
    \label{fig:snr_match_kodak}
    \end{subfigure}
    \caption{Performance of BPG-LDPC-QAM, One-shot and ISEC Deep JSCC on CPP=1/6 when $SNR_{\text{train}}=SNR_{\text{test}}$. }
    \label{fig:snr_match}
\end{figure*}

\textbf{Dataset and Evaluation Metrics}
We conduct experiments to compare ISEC with one-shot decoding on a high resolution Kodak\footnote{\url{http://r0k.us/graphics/kodak/}} dataset, which is composed of 24 $768\times 512$ pixel color images, as well as CIFAR-10 \cite{krizhevsky2009learning} low resolution image dataset, which contains $60,000$ $32\times 32$ pixel color images.
To quantify the distortion between $\vx$ and $\hat{\vx}$, we choose peak signal-to-noise ratio (PSNR),
    $PSNR=10\cdot \log_{10}\frac{255^2}{MSE}, \: MSE=\frac{\|\vx-\hat{\vx}\|_2^2}{\inpd},$
as well as structural similarity (SSIM, for CIFAR-10) \cite{ssim}, and  multi-scale version of SSIM (MS-SSIM, for Kodak) \cite{wang2003multiscale}.
For the perceptual quality we use LPIPS \cite{zhang2018unreasonable} distance and Fréchet inception distance (FID) \cite{heusel2017gans}, which is based on the distance between features of images processed by a convolutional neural network (CNN).
LPIPS compares the instance-wise difference of the deep features from VGGNet \cite{simonyan2014very}, and FID captures the distributional difference of deep features from InceptionV3 \cite{szegedy2016rethinking} network between the set of original source images and the set of reconstructed images.

\begin{figure}
    \centering
    \includegraphics[width=1.0\linewidth]{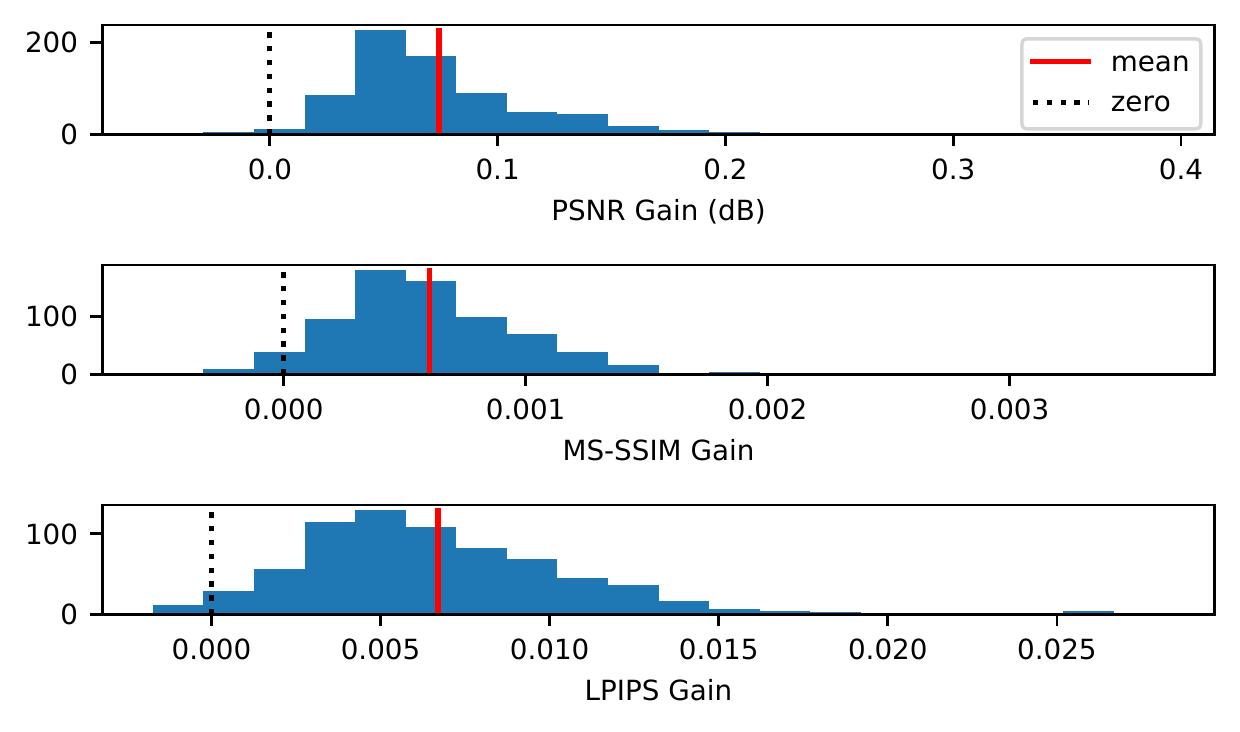}
    \caption{Histograms of ISEC gains for randomly cropped $256\times 256$ Kodak images using 7dB model, CPP=1/16.}
    \label{fig:kodak_hist}
\end{figure}



\textbf{Deep JSCC Encoder, Decoder, and Denoiser Structures}
We parameterize the Deep JSCC encoders and decoders using fully-convolutional neural networks (CNNs) by following the approach in \cite{bourtsoulatze2019deep} which allows us to encode and decode an image regardless of the image size.
Also, to ensure the injectivity of the decoder network, we use a parameterized rectified linear unit (PReLU) \cite{he2015delving} as the activation function.
Finally, the output of the encoder is normalized to have a maximum norm of $\sqrt{k}$ to satisfy the boundedness of the codeword.
Specific configurations of the networks are presented in Appendix \ref{sec:append_network_conf_deep_jscc}.

We use a bias-free version of convolutional neural network (CNN) denoiser \cite{mohan2020robust} adopting the structure in \cite{kadkhodaie2020solving} as specified in Appendix \ref{sec:append_network_conf_denoiser}.

\textbf{Training and Testing}  
 We used Open Images V6 \cite{OpenImages2} and the CIFAR-10 training dataset to train Deep JSCC encoders and decoders for evaluating Kodak and CIFAR-10 test datasets, respectively.
 We evaluate the low compression ratio models with CPP=1/6 for both CIFAR-10 and Kodak datasets, and high compression ratio models with CPP=1/12 and 1/16 for CIFAR-10 and Kodak datasets, respectively.
We use the term an \textit{$n$-dB model} to denote the Deep JSCC encoder and decoder pair trained with $n$-dB additive white Gaussian noise channel SNR.
See \ref{sec:append_network_conf_deep_jscc} and \ref{sec:append_training} for details.

\textbf{Parameters:}
We run ISEC for $T=50$ iterations for all our experiments.
For the Kodak dataset, we report the quality and distortion metrics by averaging metric values from 10 independent noise realizations for each image to reduce the variance caused by the random channel noise.
Values of $\alpha, \delta,$ and $\eta$ used in the experiments are presented in Appendix \ref{sec:parameters_for_isec}.

\begin{figure}
    \centering
    \includegraphics[width=1.0\linewidth]{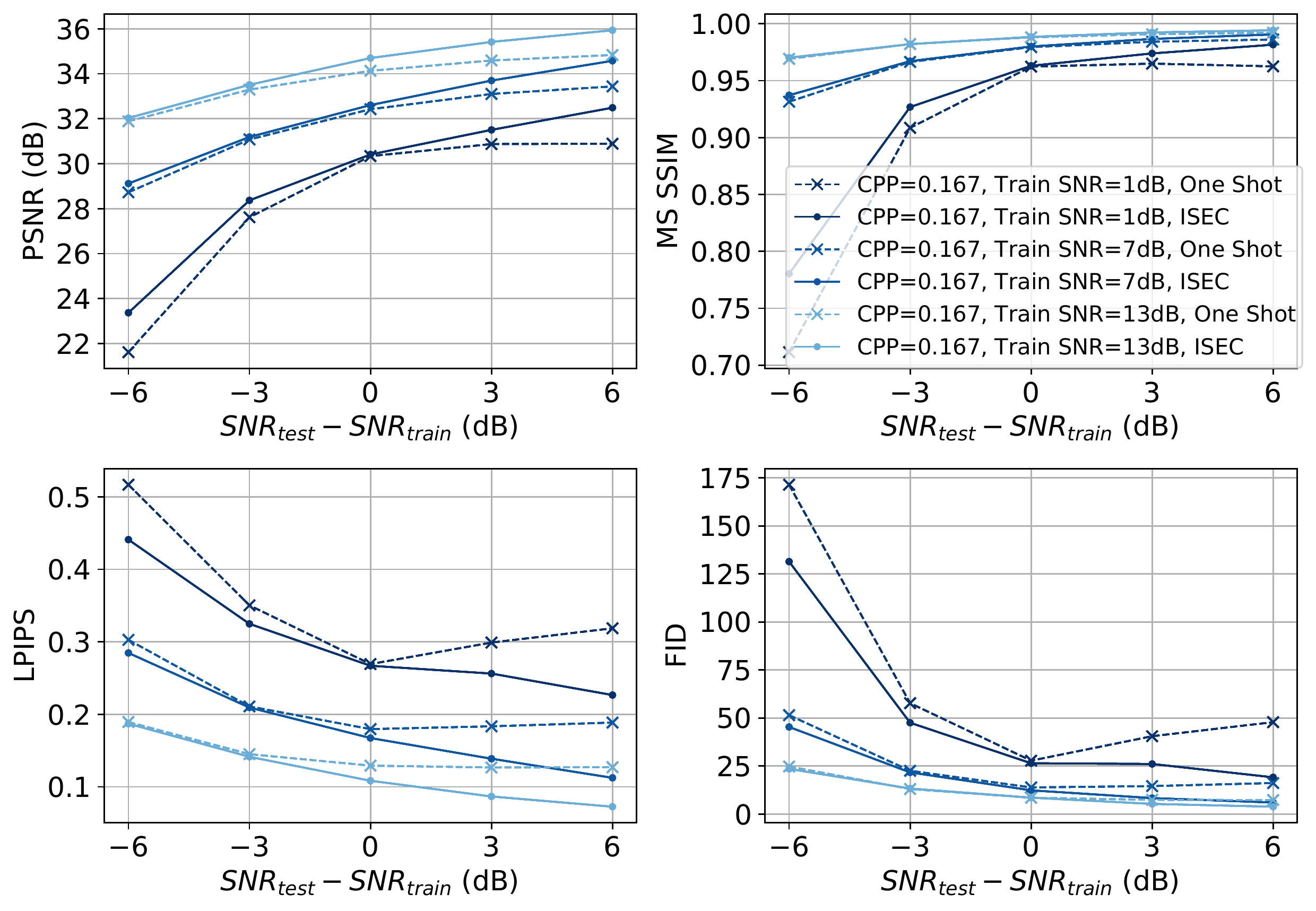}
    \caption{Reconstruction performance of ISEC on Kodak dataset when $SNR_{\text{test}} \neq SNR_{\text{train}}$, CPP=1/6. The x-axis indicates the SNR difference between the tested and trained environment. }
    \label{fig:snr_mismatch}
\end{figure}


\begin{figure*}[]
\newcommand{\ratio}{0.3}
\centering
    \begin{subfigure}{\ratio\linewidth}
    \centering
    \includegraphics[width=\linewidth]{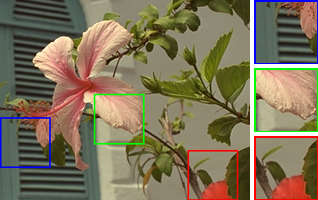}
    \caption{Target}
    \end{subfigure}
    \begin{subfigure}{\ratio\linewidth}
    \centering
    \includegraphics[width=\linewidth]{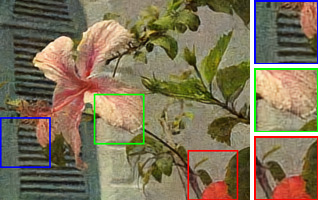}
    \caption{Before ISEC, $SNR_{\text{test}}=1$dB}
    \end{subfigure}
    \begin{subfigure}{\ratio\linewidth}
    \centering
    \includegraphics[width=\linewidth]{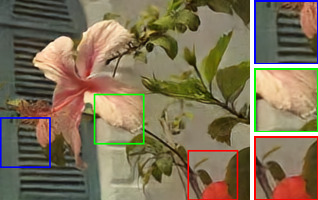}
    \caption{After ISEC, $SNR_{\text{test}}=1$dB}
    \end{subfigure}
    
        \begin{subfigure}{\ratio\linewidth}
    \centering
    \includegraphics[width=\linewidth]{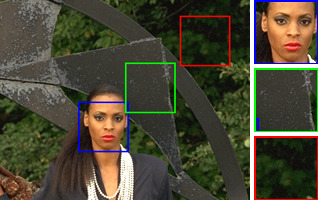}
    \caption{Target}
    \end{subfigure}
    \begin{subfigure}{\ratio\linewidth}
    \centering
    \includegraphics[width=\linewidth]{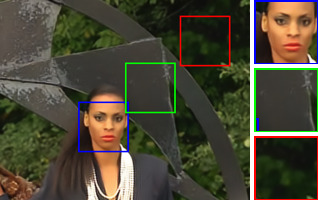}
    \caption{Before ISEC, $SNR_{\text{test}}=13$dB}
    \end{subfigure}
    \begin{subfigure}{\ratio\linewidth}
    \centering
    \includegraphics[width=\linewidth]{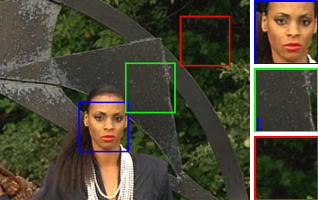}
    \caption{After ISEC, $SNR_{\text{test}}=13$dB}
    \end{subfigure}
    \caption{Before and after ISEC on Kodak dataset using $SNR_{\text{train}}$=7dB model on CPP=1/16. All images are cropped.}
    \label{fig:mismatch_visual}
\end{figure*}

\subsection{Performance Under Trained Environment}

We first examine our algorithm when the distribution of the channel noise is not different from that used during the training of the Deep JSCC model.
The PSNR, SSIM, LPIPS, and FID metrics before and after ISEC for CIFAR-10 and Kodak images with CPP=1/6 are presented in Figure \ref{fig:snr_match}. 
Note that higher PSNR and SSIM indicate better quality while lower LPIPS and FID are desired.
For the classical non-learning-based baseline, we include Better Portable Graphics (BPG) \cite{bellad2018bpg} and low-density parity-check (LDPC) code \cite{gallager1962low} with BPSK,  4-, or 16-QAM modulation.
No encoding and decoding failure was observed when CPP=1/6 for the evaluation in Figure \ref{fig:snr_match}.
BPG encoding or decoding failures can happen in CPP=1/12 or 1/16 settings. We fill the reconstruction with the mean of the input image when BPG encoding fails. Similarly, we use a zero-valued image for reconstruction if BPG decoding fails.

For CIFAR-10 images, ISEC shows moderate gain in all 0, 5, and 10dB SNR cases compared to one-shot JSCC.
We suppose the gain is moderate because the dimension of the images in the dataset is small so the JSCC encoder and decoder pair learns near-optimal coding and decoding for the trained SNR. On the contrary, the gains are more significant as shown in Figure \ref{fig:snr_match_kodak} when ISEC is applied to the high-resolution Kodak image dataset.

As reported in \cite{bourtsoulatze2019deep}, the conventional (separate) source-channel coding scheme (BPG-LDPC-QAM) outperforms Deep JSCC (and ISEC) schemes on PSNR (for high-resolution images such as Kodak).
However, the perceptual metrics (MS-SSIM, LPIPS, and FID) are better with Deep JSCC and ISEC especially when SNR is low or moderate.
The same trend was observed when CPP=1/16 as presented in Appendix.

Figure \ref{fig:kodak_hist} shows the distribution of the gains from ISEC on the Kodak images evaluated with the 7dB model at CPP=1/16.
Here we randomly sample 30 patches of $256\times 256$ pixels from each Kodak image.
We do not include FID for the histogram because it is measured on the entire set of images. 
Histograms show most patches exhibit a positive gain in all three metrics. 
The same tendency is observed from 1dB and 13dB models as well in Appendix.

\begin{figure}[]
    \centering
    \includegraphics[width=1.0\linewidth]{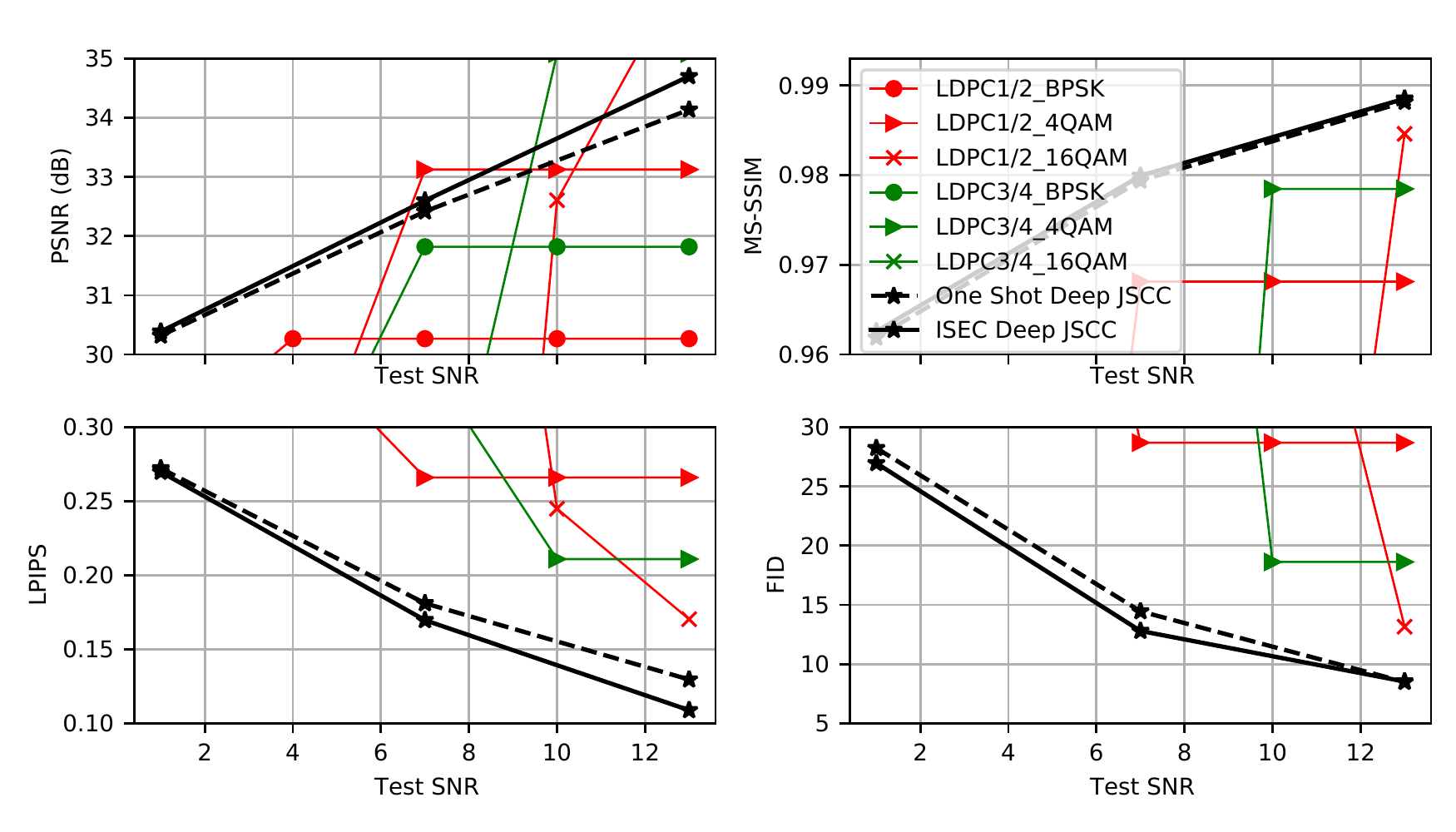}
    \caption{Average performance of Kodak images under additive white \textit{Laplacian} noise with CPP=1/6 using the models trained with AWGN. ISEC generalizes well for the out-of-distribution noise.}
    \label{fig:kodak_laplace}
\end{figure}

\begin{figure*}[]
    \centering
    \includegraphics[width=1.0\linewidth]{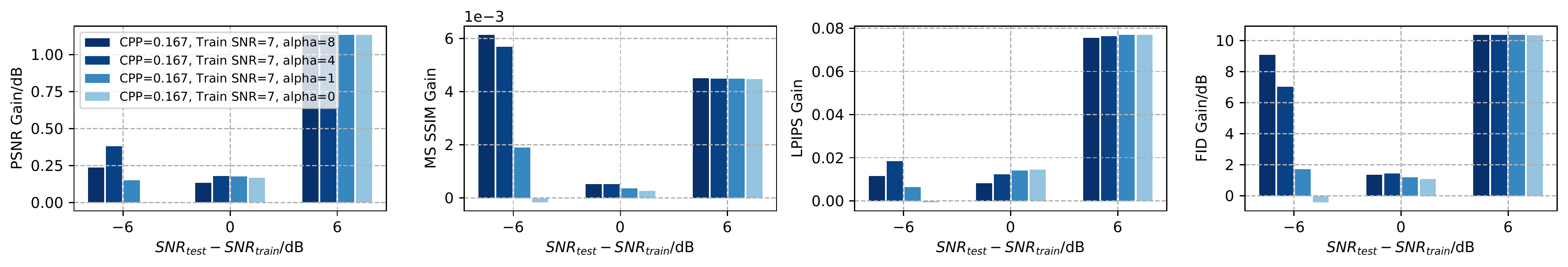}
    \caption{Quality metric gains observed from different $\alpha\in\{8, 4, 1, 0\}$. }
    \label{fig:kodak_alpha_avg}
\end{figure*}

\subsection{Performance Under Mismatched SNRs}

We evaluate ISEC when the test SNR is different from the trained SNR. It represents a practical scenario where the transmitter does not have the knowledge of the channel SNR which is estimated at the receiver after receiving/observing the noisy codeword. 
As shown in Figure \ref{fig:snr_mismatch}, ISEC successfully handles mismatched test SNRs when they are higher or lower than the trained SNR when CPP=1/6. The results for CPP=1/16 are provided in Appendix. 
Note that, unlike Deep JSCC and ISEC, the conventional BPG-LDPC-QAM scheme experiences catastrophic decoding failures  (image is not decodable) when the test/actual SNR is lower than targeted, whereas the receiver cannot improve the reconstruction quality when the test SNR is higher than targeted.

We show visual comparisons in Figure \ref{fig:mismatch_visual} using example Kodak images (more examples in Appendix) when the test SNR is 6dB lower than the trained SNR. 
It is observed that a one-shot Deep JSCC decoder adds many artifacts whereas ISEC can \textit{correct} some of them to improve the quality. 
\textit{Implicit prior} captured by the denoiser network plays a substantial role when the test  SNR deteriorates from the training SNR, which is further analyzed in Section \ref{sec:impact_of_prior}.
Intuitively speaking, when $SNR_{\text{test}}<SNR_{\text{train}}$, the gradient of the log likelihood in \eqref{eq:ISEC_Likelihood} can be in the wrong direction due to the severe noise.
However, this can be compensated by the denoiser network which estimates the gradient of the log prior predicting the (rough) direction toward the original codeword through the iterative update process.

It is worth noting that Figure \ref{fig:snr_mismatch} and \ref{fig:mismatch_visual} show that the reconstruction quality of the one-shot JSCC remains almost the same or even degrades producing blurry images when the test SNR is higher than training SNR. 
On the other hand, ISEC successfully recovers the details from the blurry initial (one-shot) reconstruction.

Although our framework uses a non-linear encoder $(\vy=\enc(\vx)+\vn)$ unlike linear measurements  in CSGM $(\vy=\mA\vx+\vn)$, the gain observed by ISEC in high SNRs is consistent with the claim in the CSGM literature \cite[Thm. 1.2.]{bora2017compressed} that the reconstruction error is proportional to the norm of the observation noise and the number of measurements. 
However, for general rate-quality tradeoffs, it is known that Deep JSCC (even without ISEC) outperforms CSGM which relies on a linear encoder $\mA$ \cite{grover2019uncertainty}. 
In summary, ISEC enhances the robustness of Deep JSCC against unexpected downward SNR whereas the reconstruction quality improves by ISEC when the actual SNR is higher than expected. 
The same benefits are unattainable in conventional separate source-channel coding schemes.

\subsection{Performance in Additive White Laplace Noise}

In this section, we evaluate the case where the noise distribution is shifted from the trained.
We test the Kodak dataset in an \textit{additive white Laplace noise} (AWLN) channel with CPP=1/6 that has the same variance as the Gaussian noise (AWGN) channel used during the training of Deep JSCC models $(\enc, \dec)$ and the denoiser network $F$.
The same decoding algorithm and parameter selection rules specified in Algorithm \ref{alg:ISEC_basic} are applied to evaluate the AWLN case.
Figure \ref{fig:kodak_laplace} shows that ISEC achieves performance improvement from one-shot decoding in this noise distribution mismatch scenario. 
Despite the same SNR, the conventional BPG-LDPC-QAM scheme undergoes performance degradation under the AWLN channels compared to the AWGN case, showing worse perceptual quality metrics (MS-SSIM, LPIPS, and FID) than the Deep JSCC one-shot decoding and ISEC.
On the other hand, after applying ISEC, all evaluation criteria are improved from the one-shot decoding despite being performed under white Laplace distribution. 
Note that neither transmitters nor receivers are aware of the distributional shift of the channel noise at any stage.
This shows that our proposed decoding algorithm generalizes well and enhances the robustness of the Deep JSCC scheme under the unseen channel noise distribution.

\begin{figure}[]
    \centering
    \includegraphics[width=1.0\linewidth]{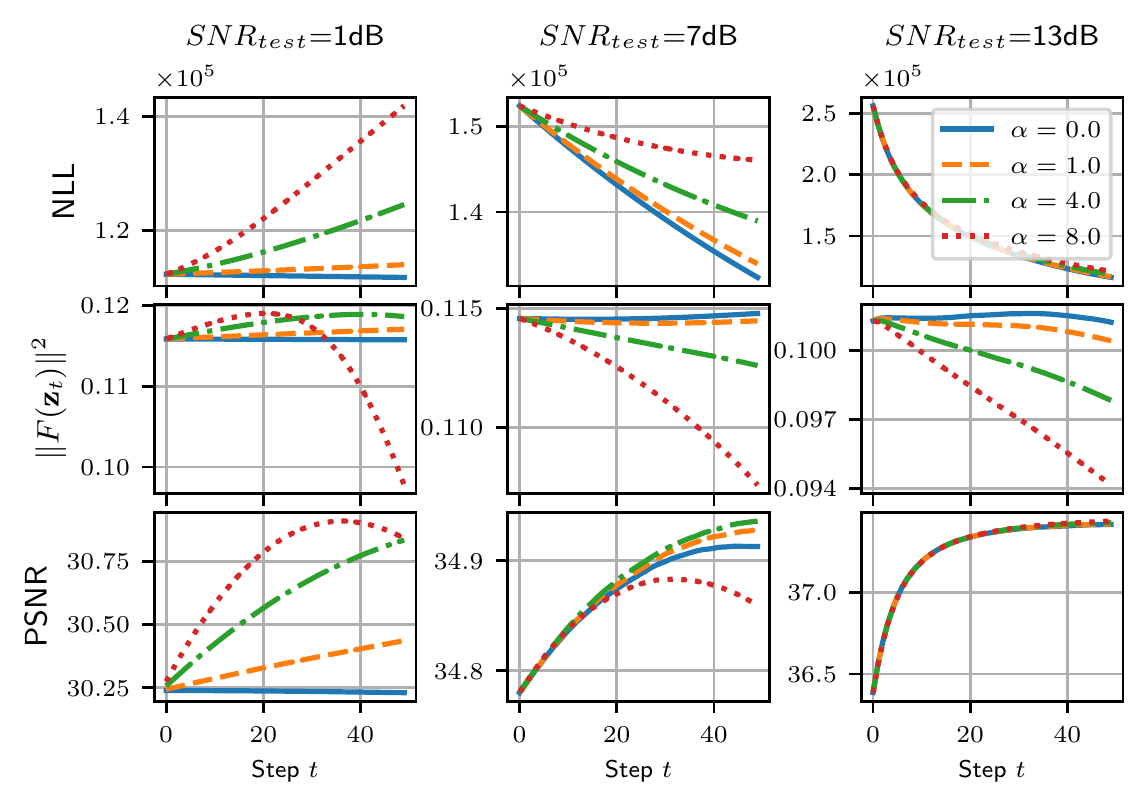}
    \caption{ISEC objectives over steps using the 7dB Deep JSCC model with CPP=1/6 on \texttt{kodim06}.}
    \label{fig:kodim06_alphas_st_7_s_1}
\end{figure}

\subsection{Role of Implicit Prior}\label{sec:impact_of_prior}

We study the role of the implicit prior in ISEC with varying $\alpha$ in \eqref{eq:GD_Final} and \eqref{eq:alpha_eta} which scales the output of the denoiser to approximate the gradient of the log prior density of the codeword.
A larger $\alpha$ makes each step rely more on the prior information.
We tested the 7dB model with CPP=1/6 in 1, 7, and 13dB SNRs with $\alpha\in\{0, 1, 4, 8\}$ for 3 different test SNRs.  Figure \ref{fig:kodak_alpha_avg} summarizes the results.
When the test SNR is 13dB, the $\alpha$ does not affect the performance much. 
However, higher gains are achieved by using a proper parameter $\alpha$ when the test SNR is the same or lower than the training SNR. It indicates that the approximated gradient of log prior density plays a key role when the observed SNR is lower than that of the training environment.

We also show in Figure \ref{fig:kodim06_alphas_st_7_s_1} how the negative likelihood (NLL) $\frac{1}{2\sigma^2}\|\vy-\enc(\dec(\vz_t))\|_2^2$, the squared norm of the denoiser output $\|F(\vz_t)\|_2^2$, and the PSNR of the reconstruction change along the ISEC step $t=0,\ldots,49$ with different $\alpha$'s.
When the test SNR is 1dB, notice the PSNR does not improve by ISEC with $\alpha=0$ and $1$, whereas when $\alpha=4$, all metrics improve from one-shot decoding through ISEC iterations.
We also note that when $\alpha=4$ or $8$, the NLL  increases from the starting point as PSNR enhances while the squared norm of the denoiser rapidly decreases (i.e., low $\|\nabla\log p(\vz_t)\|^2$).
This shows that the denoiser with the proper choice of $\alpha$ can deal with noise variance higher than the trained environment because of the bias-free structure.
When $\alpha=8$, however, the prior information dominates the likelihood too much so that the PSNR drops after a few tens of iterations.
As the test SNR increases to 13dB, the impact of the denoiser becomes negligible and the likelihood dominates. 


\section{CONCLUSION}

We introduced ISEC for Deep JSCC for the scenario employing a power-limited transmitter and a resource-abundant receiver.
Our method extends the Deep JSCC decoding in an iterative way by correcting the source error in the codeword space.
We introduce a modified MAP decoding algorithm with an approximated gradient of the log density using a bias-free denoiser network. Through iterations, the missed information from the one-shot decoding can be recovered by ISEC to improve various quality measures, especially when the observed channel noise characteristics are mismatched from the training environment.

\textbf{Limitations and Future Work}. 
Although we assume the receiver has abundant computing resources unlike the resource-constrained transmitter, the iterative decoding in ISEC increases the computational cost and power consumption compared to the one-shot decoding.
Reducing the number of iterations to lower the complexity of ISEC is a possible extension of this work. Applying and evaluating ISEC to different source types such as audio and sensor data is left as future work.

\subsubsection*{Acknowledgments}
This research was supported in part by NSF CAREER Award \#1942806,
and also
through computational resources and services provided by Advanced Research Computing at the University of Michigan, Ann Arbor.

\bibliographystyle{plainnat}
\bibliography{sec_bib}

\clearpage

\appendix
\onecolumn



\section{Network Configurations, Training, and Test Details}

This Appendix section presents the configurations and training methods of deep neural networks used in our paper.
The parameters of ISEC used in Section \ref{sec:experiments} are also specified.

\subsection{Deep JSCC Encoder and Decoder}\label{sec:append_network_conf_deep_jscc}

\subsubsection{Configuration}
Deep JSCC encoders and decoders are convolutional neural networks \cite{krizhevsky2012imagenet,simonyan2014very,he2016deep}.
In both networks, batch normalization \cite{pmlr-v37-ioffe15} is used between each convolution layer and a non-linear activation.
Specifically, we use the following parameterized rectified linear unit (PReLU) \cite{he2015delving} as the activation function
\begin{align*}
    \mathsf{PReLU}(u) = \begin{cases}
        u & u \ge 0 \\
        \rho \cdot u & u < 0 
    \end{cases}
\end{align*}
where $\rho\ge0$ is a trainable parameter shared within the layer.
We use convolution layers with stride 2 for downsampling, and bilinear upsampling with a factor of 2 for upsampling the width and height of the tensor at the encoder and decoder, respectively.
Hence the encoder network downsamples the input image with size $\inpd = H\times W \times 3$ into $\coded = \frac{HWC}{4\times 4}$, and the decoder network maps the compressed (noisy) codeword back into the original pixel domain.
Also, the power normalization layer is added at the end of the encoder to satisfy the power constraint $\|\enc(\vx)\|_2^2 \le \coded$:
\begin{align*}
\text{PowerNorm}(\vv, k) = \begin{cases}
        \vv & \text{if } \|\vv\|_2^2 \le k \\
        \sqrt{k}\frac{\vv}{\|\vv\|_2} & \text{otherwise}
    \end{cases}.
\end{align*}
Residual blocks \cite{he2016deep} depicted in Figure \ref{fig:residual} are added to the network for the overall configuration specified in Table \ref{tab:enc_dec_structure_cifar10} and \ref{tab:enc_dec_structure_kodak} for CIFAR-10 and Kodak datasets, respectively.

\subsubsection{Training Method}
The weights and biases of Deep JSCC encoders and decoders are updated by minimizing the mean squared error (i.e., \eqref{eq:deep_jscc_training}) between the input $\vx$ and the reconstruction from the noisy codeword $\dec(\enc(\vx)+\noise)$, where $\noise$ is a channel noise.

For low-resolution images, Deep JSCC encoders and decoders are trained with CIFAR-10 \cite{krizhevsky2009learning} training set, which contains $50,000$ $32\times 32$ color images of 10 different classes. 
We use the batch size of 64 for 234,300 steps with Adam \cite{kingma2014adam} optimizer with parameter $\beta_1 = 0.0$ and $\beta_2=0.9$.
The learning rate was set to 0.0002 and reduced to 0.00002 at the 117,150th step.

For high-resolution images, we use a subset of 273,230 images from Open Images V6 \cite{OpenImages2} training dataset\footnote{
We use the first two tarball file (\texttt{train\_0.tar.gz} and \texttt{train\_1.tar.gz}) of the training set from the following website: {\texttt{https://github.com/cvdfoundation/open-images-dataset}}}.
The images are randomly cropped to $128\times 128$ pixel patches.
We use the batch size of 128 and train the model for 96,030 steps with Adam \cite{kingma2014adam} optimizer with parameter $\beta_1 = 0.0$ and $\beta_2=0.9$, where the learning rate is set to 0.001 and reduced to 0.0001 at the 64,020th step.

We set CPP=1/6 or 1/12 for low-resolution images, and CPP=1/6 or 1/16 for high-resolution images.
For each CPP evaluation, we train three encoder-decoder pairs for three channel SNR cases of 0, 5, and 10 dB for low-resolution images. Similarly, three pairs are trained for 1, 7, and 13 dB SNR for high-resolution images per each CPP. 
We use the term \textit{$n$-dB model} to denote the Deep JSCC encoder and decoder pair trained with $n$-dB channel noise.
All networks are trained on a 2.9 GHz Intel Xeon Gold 6226R processor and one NVIDIA A40 GPU.

\subsection{Bias-Free Denoiser}\label{sec:append_network_conf_denoiser}

\subsubsection{Configuration}
\begin{figure}[h]
    \centering
    \subcaptionbox{\label{fig:residual}}[0.4\linewidth]{
     \includegraphics[width=0.7\linewidth]{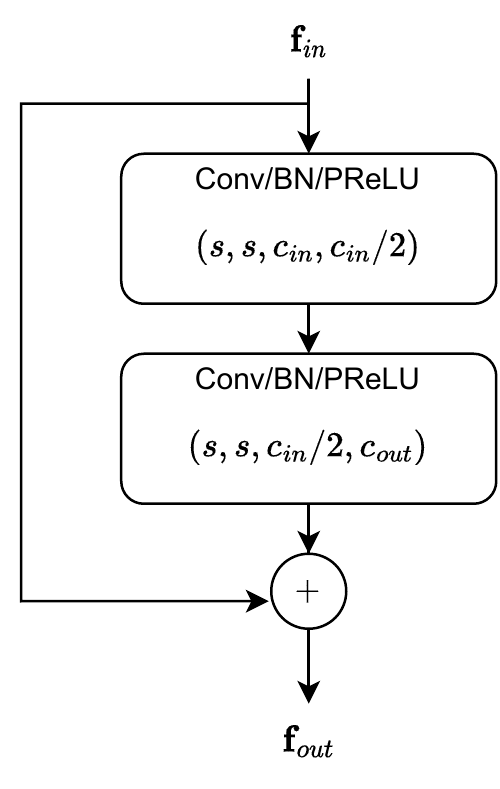}
     }
    \subcaptionbox{\label{fig:bfcnn}}[0.4\linewidth]{
     \includegraphics[width=0.7\linewidth]{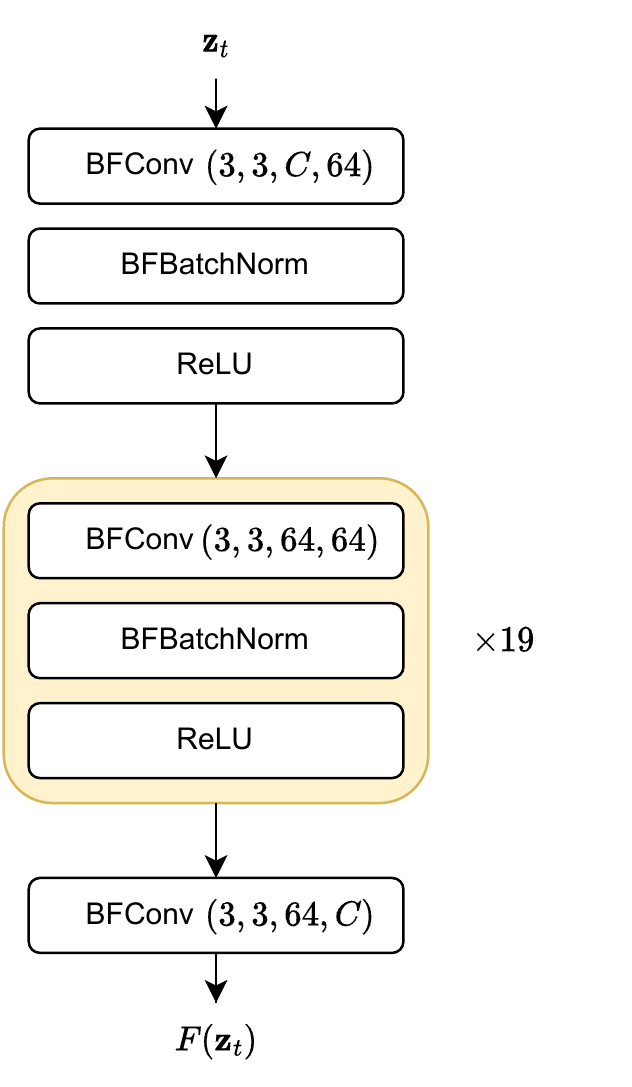}}
    \caption{(a): Residual Block used in Deep JSCC encoders and decoders. The elements of the tuple $(s,s,c,d)$ denote the width and height of the kernel, and the numbers of input and output channels of the convolution layer, respectively. If $c_{in}\neq c_{out}$, a $1\times 1$ convolution layer is applied to the shortcut. (b): Bias-free Denoiser.}
\end{figure}

For the codeword (latent) denoiser, we use a bias-free denoiser \cite{mohan2020robust}.
Specifically, we use 20 Convolution-Batch Normalization-ReLU blocks \textit{without any additive bias terms} based on the structure in \cite{kadkhodaie2020solving}. 
As illustrated in Figure \ref{fig:bfcnn}, the bias-free convolution layers are composed of $3\times 3\times 64$ kernels except for the last layer.

\subsubsection{Training Method}\label{sec:append_training}

Bias-free denoisers used in this paper operate in the \textit{codeword} domain.
Thus we trained each denoiser (there are multiple versions with different latent/codeword sizes) to predict the \textit{channel noise instance} from a corrupted codeword.
That is, given $\vx_i$ in the training set we first map $\vx_i$ to the codeword $\enc(\vx_i)$, where $\enc:\R^\inpd \to \R^\coded$ is a Deep JSCC encoder network.
Then we sample a $\coded$-dimensional channel noise following the predefined noise distribution $\noise_i \sim P_{\sigma}(N)$ which does not change during the training. Then, we update the weights of the denoiser network $F':\R^\coded \to \R^\coded$ to minimize the mean squared error between the noise instance $\noise_i$ and the output of the denoiser network $F'(\enc(\vx_i)+\noise_i)$ given the corrupted codeword $\enc(\vx_i)+\noise_i$:
\begin{align*}
    F &:=\argmin_{F'} \frac{1}{\nptstrain}\sum_{i=1}^{\nptstrain}\|\noise_i - F'(\enc(\vx_i)+\noise_i)\|_2^2.
\end{align*}

The bias-free denoisers for CIFAR-10 are trained using codewords of the CIFAR-10 training data.
The models are trained for 234,300 steps with batch size of 64 using Adam optimizers with $\beta_1=0.9$ and $\beta_2=0.999$.
The learning rate is set to 0.0002 and decayed to 0.00002 at the 117,150th step.

The bias-free denoisers for Kodak experiments are trained with codewords of randomly cropped $128\times 128$ pixel patches in the subset of the Open Images V6 training dataset used for Deep JSCC model training. 
We trained each bias-free denoiser for 53,350 steps with batch size of 128 using Adam optimizer with $\beta_1=0.9$ and $\beta_2=0.999$.
We kept the learning rate policy the same as the CIFAR-10 denoiser training.

\begin{table}[p]
\centering
\begin{tabular}{|lccc|}
\hline
\multicolumn{1}{|c|}{Type}                & \multicolumn{1}{c|}{Kernel Size ($s$)} & \multicolumn{1}{c|}{Stride$\downarrow$, Upscale$\uparrow$} & Output Shape ($C_{h}\times H\times W$) \\ \hline
\multicolumn{4}{|c|}{Encoder}                                                                         \\ \hline
\multicolumn{1}{|l|}{Conv/BN/PReLU} & \multicolumn{1}{c|}{7} & \multicolumn{1}{c|}{1} & (32, 32, 32)  \\ \hline
\multicolumn{1}{|l|}{Conv/BN/PReLU} & \multicolumn{1}{c|}{5} & \multicolumn{1}{c|}{2$\downarrow$} & (64, 16, 16)  \\ \hline
\multicolumn{1}{|l|}{Residual}      & \multicolumn{1}{c|}{3} & \multicolumn{1}{c|}{1} & (64, 16, 16)  \\ \hline
\multicolumn{1}{|l|}{Conv/BN/PReLU} & \multicolumn{1}{c|}{5} & \multicolumn{1}{c|}{2$\downarrow$} & (128, 8, 8)   \\ \hline
\multicolumn{1}{|l|}{Residual}      & \multicolumn{1}{c|}{3} & \multicolumn{1}{c|}{1} & (128, 8, 8)   \\ \hline
\multicolumn{1}{|l|}{Residual}      & \multicolumn{1}{c|}{3} & \multicolumn{1}{c|}{1} & (128, 8, 8)   \\ \hline
\multicolumn{1}{|l|}{Residual}      & \multicolumn{1}{c|}{3} & \multicolumn{1}{c|}{1} & (128, 8, 8)   \\ \hline
\multicolumn{1}{|l|}{Residual}      & \multicolumn{1}{c|}{3} & \multicolumn{1}{c|}{1} & ($\coded$/64, 8, 8)   \\ \hline
\multicolumn{1}{|l|}{Power Normalization} & \multicolumn{1}{c|}{-}           & \multicolumn{1}{c|}{-}            & ($\coded$/64, 8, 8)                            \\ \hline
\multicolumn{4}{|c|}{Decoder}                                                                         \\ \hline
\multicolumn{1}{|l|}{Conv/BN/PReLU} & \multicolumn{1}{c|}{3} & \multicolumn{1}{c|}{1} & (128, 8, 8)   \\ \hline
\multicolumn{1}{|l|}{Residual}      & \multicolumn{1}{c|}{3} & \multicolumn{1}{c|}{1} & (128, 8, 8)   \\ \hline
\multicolumn{1}{|l|}{Residual}      & \multicolumn{1}{c|}{3} & \multicolumn{1}{c|}{1} & (128, 8, 8)   \\ \hline
\multicolumn{1}{|l|}{Upsample}      & \multicolumn{1}{c|}{-} & \multicolumn{1}{c|}{2$\uparrow$} & (128, 16, 16) \\ \hline
\multicolumn{1}{|l|}{Conv/BN/PReLU} & \multicolumn{1}{c|}{3} & \multicolumn{1}{c|}{1} & (64, 16, 16)  \\ \hline
\multicolumn{1}{|l|}{Residual}      & \multicolumn{1}{c|}{3} & \multicolumn{1}{c|}{1} & (64, 16, 16)  \\ \hline
\multicolumn{1}{|l|}{Upsample}      & \multicolumn{1}{c|}{-} & \multicolumn{1}{c|}{2$\uparrow$} & (64, 32, 32)  \\ \hline
\multicolumn{1}{|l|}{Conv/BN/PReLU} & \multicolumn{1}{c|}{3} & \multicolumn{1}{c|}{1} & (32, 32, 32)  \\ \hline
\multicolumn{1}{|l|}{Residual}      & \multicolumn{1}{c|}{3} & \multicolumn{1}{c|}{1} & (32, 32, 32)  \\ \hline
\multicolumn{1}{|l|}{Conv/BN/Tanh}  & \multicolumn{1}{c|}{5} & \multicolumn{1}{c|}{1} & (3, 32, 32)   \\ \hline
\end{tabular}
\caption{Deep JSCC Encoder and Decoder structure for CIFAR-10. Kernel size: the width and height of the kernel of the convolution layer, $C_h$: number of hidden channels of the network, $\coded$: dimension of the codeword space. }
\label{tab:enc_dec_structure_cifar10}
\end{table}

\begin{table}[p]
\centering
\begin{tabular}{|lccc|}
\hline
\multicolumn{1}{|c|}{Type}                & \multicolumn{1}{c|}{Kernel Size ($s$)} & \multicolumn{1}{c|}{Stride$\downarrow$, Upscale$\uparrow$} & Output Shape ($C_{h}\times H\times W$) \\ \hline
\multicolumn{4}{|c|}{Encoder}                                                                             \\ \hline
\multicolumn{1}{|l|}{Conv/BN/PReLU} & \multicolumn{1}{c|}{7} & \multicolumn{1}{c|}{2$\downarrow$} & $(128, H/2, W/2)$ \\ \hline
\multicolumn{1}{|l|}{Conv/BN/PReLU} & \multicolumn{1}{c|}{5} & \multicolumn{1}{c|}{2$\downarrow$} & $(128, H/4, W/4)$ \\ \hline
\multicolumn{1}{|l|}{Residual}      & \multicolumn{1}{c|}{5} & \multicolumn{1}{c|}{1} & $(128, H/4, W/4)$ \\ \hline
\multicolumn{1}{|l|}{Residual}      & \multicolumn{1}{c|}{5} & \multicolumn{1}{c|}{1} & $(128, H/4, W/4)$ \\ \hline
\multicolumn{1}{|l|}{Residual}      & \multicolumn{1}{c|}{5} & \multicolumn{1}{c|}{1} & $(128, H/4, W/4)$ \\ \hline
\multicolumn{1}{|l|}{Conv/BN}       & \multicolumn{1}{c|}{5} & \multicolumn{1}{c|}{1} & $(\frac{16\coded}{HW}, H/4, W/4)$   \\ \hline
\multicolumn{1}{|l|}{Power Normalization} & \multicolumn{1}{c|}{-}           & \multicolumn{1}{c|}{-}            & $(\frac{16\coded}{HW}, H/4, W/4)$                        \\ \hline
\multicolumn{4}{|c|}{Decoder}                                                                             \\ \hline
\multicolumn{1}{|l|}{Conv/BN/PReLU} & \multicolumn{1}{c|}{5} & \multicolumn{1}{c|}{1} & $(128, H/4, W/4)$ \\ \hline
\multicolumn{1}{|l|}{Residual}      & \multicolumn{1}{c|}{5} & \multicolumn{1}{c|}{1} & $(128, H/4, W/4)$ \\ \hline
\multicolumn{1}{|l|}{Residual}      & \multicolumn{1}{c|}{5} & \multicolumn{1}{c|}{1} & $(128, H/4, W/4)$ \\ \hline
\multicolumn{1}{|l|}{\textit{Upsample}}      & \multicolumn{1}{c|}{-} & \multicolumn{1}{c|}{2$\uparrow$} & $(128, H/2, W/2)$ \\ \hline
\multicolumn{1}{|l|}{Conv/BN/PReLU} & \multicolumn{1}{c|}{5} & \multicolumn{1}{c|}{1} & $(128, H/2, W/2)$ \\ \hline
\multicolumn{1}{|l|}{Residual}      & \multicolumn{1}{c|}{5} & \multicolumn{1}{c|}{1} & $(128, H/2, W/2)$ \\ \hline
\multicolumn{1}{|l|}{\textit{Upsample}}      & \multicolumn{1}{c|}{-} & \multicolumn{1}{c|}{2$\uparrow$} & $(128, H, W)$     \\ \hline
\multicolumn{1}{|l|}{Conv/BN/Tanh}  & \multicolumn{1}{c|}{7} & \multicolumn{1}{c|}{1} & $(3, H, W)$       \\ \hline
\end{tabular}
\caption{Deep JSCC Encoder and Decoder structure for Kodak dataset. $C_h$: number of hidden channels of the network, $\coded$: dimension of the codeword space. For \textit{Upsample}, we used bilinear upsampling.}
\label{tab:enc_dec_structure_kodak}
\end{table}

\subsection{Data pre- and post-processing}
All image pixels are normalized from the integer values $[0, 255]$ to floating point values $[-1, 1]$ before being fed into the network.
During testing, we use the output of the decoder in the range of $[-1, 1]$ directly to evaluate LPIPS and FID. 
The decoder outputs are re-mapped to the original integer range of $[0, 255]$ for the evaluation of PSNR and (MS-)SSIM.
Random horizontal flipped images were created to augment the dataset during training.

\subsection{Parameters for ISEC}\label{sec:parameters_for_isec}
In this section, we specify the parameters of Algorithm \ref{alg:ISEC_basic} used in our experiments.
Note that $\alpha$ and $\delta$ are used for calculating $\alpha'$ in Algorithm \ref{alg:ISEC_basic}, which controls the effect of the gradient of implicit prior (i.e., \eqref{eq:gradient_logprior}), whereas $\eta$ is a step size of ISEC. 
Throughout CIFAR-10 evaluations, $\alpha$ and $\delta$ are fixed to $1.0$. Regardless of CPP, we use $\eta=0.004$, $0.002$, and $0.001$ for 0, 5, and 10dB models, respectively.
For the Kodak dataset, we set $\alpha=2$ for 1dB models and $\alpha=4$ for 7 and 13dB models. 
When CPP=1/6, the parameter pair $(\delta, \eta)$ is set to $(0.5, 0.001)$, $(1.0, 0.001)$, and $(2.0, 0.005)$ for the 1, 7, and 13dB models, respectively.
When CPP=1/16, we use $(0.0, 0.001)$, $(0.0, 0.0005)$, and $(2.0, 0.005)$ for the 1, 7, and 13dB models, respectively.
In particular, we use $(\delta, \eta)=(2.0, 0.001)$ when test SNRs are 16dB and 19dB since the step size grows abnormally large otherwise.

\subsection{LDPC codes setting}
For all of our LDPC experiments on the separate source-channel coding scheme, we used the IEEE 802.11 WiFi standard with a block length of 648 and a submatrix size of 27.
The LDPC codes are decoded using the belief propagation algorithm with 10 iterations.

\begin{figure*}
    \centering
    \begin{subfigure}{\linewidth}
    \includegraphics[width=1.0\linewidth]{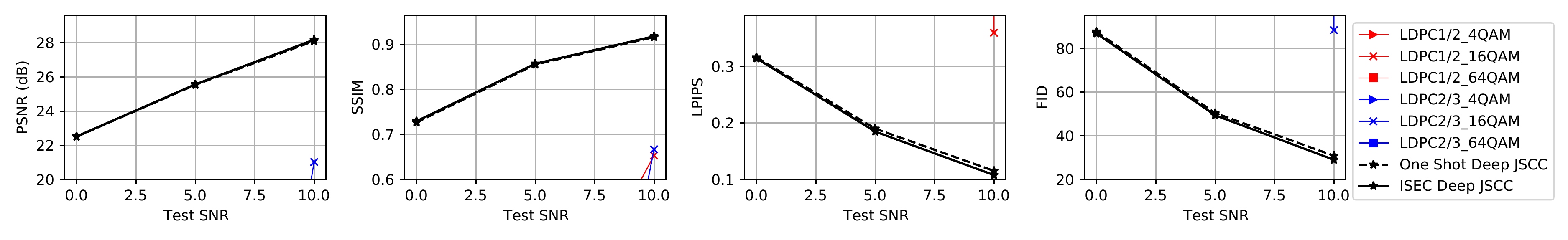}
    \caption{CIFAR-10, CPP=1/12}
    \label{fig:snr_match_cifar_lr}
    \end{subfigure}
    \begin{subfigure}{\linewidth}
    \includegraphics[width=1.0\linewidth]{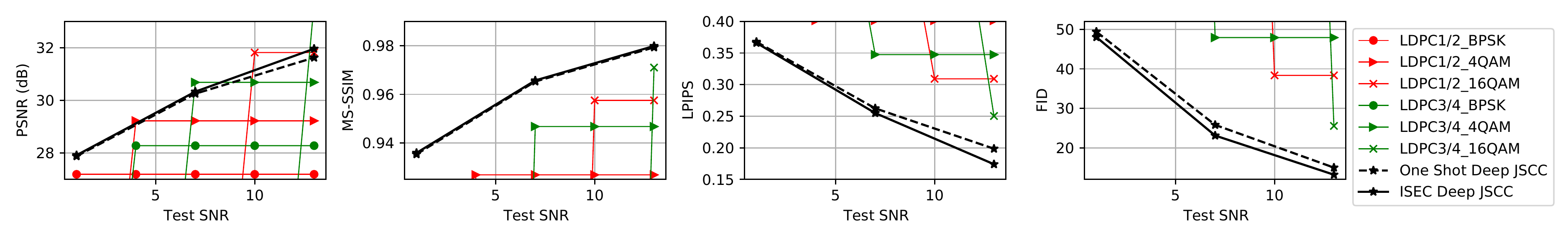}
    \caption{Kodak dataset, CPP=1/16}
    \label{fig:snr_match_kodak_lr}
    \end{subfigure}
    \caption{Performance of BPG-LDPC-QAM, One-shot and ISEC Deep JSCC when $SNR_{\text{train}}=SNR_{\text{test}}$. }
    \label{fig:snr_match_lr}
\end{figure*}

\section{Additional Quantitative Results}\label{sec:append_quantitative}

\subsection{Performance of Low Channel-per-pixel models Under Trained Environment.}

In Figure \ref{fig:snr_match_cifar_lr} and \ref{fig:snr_match_kodak_lr}, we present the performance of one-shot vs. ISEC Deep JSCC for the CIFAR-10 dataset at CPP=1/12 and the Kodak dataset at CPP=1/16, respectively. 
Compared to the performance when CPP=1/6 in Figure \ref{fig:snr_match}, ISEC and one-shot Deep JSCC decoding methods outperform BPG-LDPC-QAM with larger gaps in perceptual quality metrics.
The one-shot decoding and ISEC are tested under the same channel characteristics as the training.
ISEC increases all performance metrics in lower CPP cases although the observed gains from ISEC are smaller than those from higher CPP cases.

\subsection{Mismatched Noise Characteristics}
\begin{figure}[]
\centering
    \begin{subfigure}{\linewidth}
    \centering
        \begin{subfigure}{\linewidth}
        \centering
        \includegraphics[width=\linewidth]{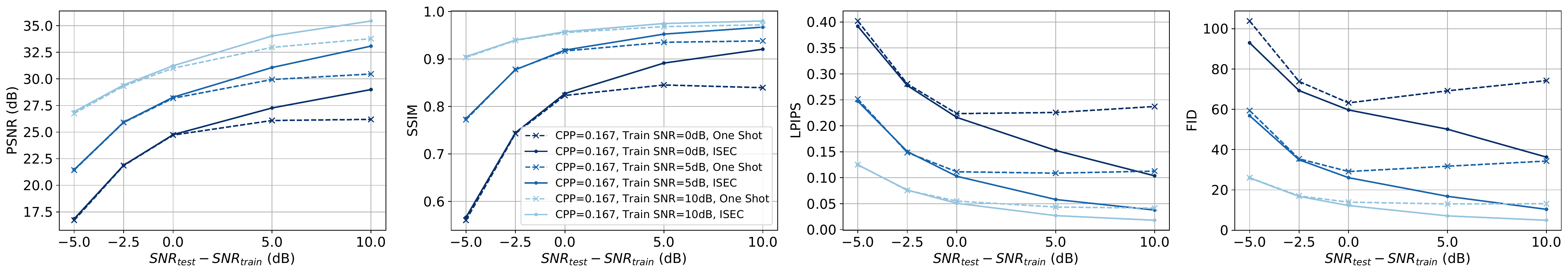}
        \end{subfigure}
        \begin{subfigure}{\linewidth}
        \centering
        \includegraphics[width=\linewidth]{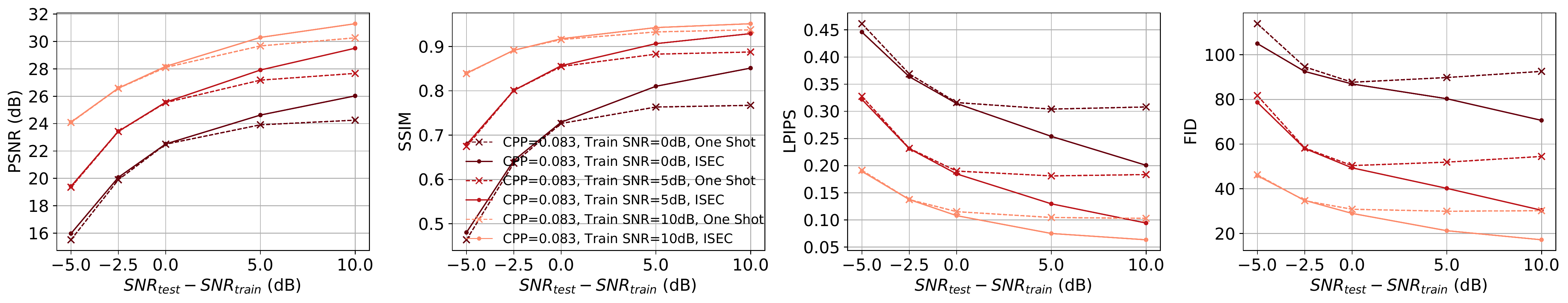}
        \end{subfigure}
    \end{subfigure}
    
         \caption{Reconstruction Performance of ISEC on CIFAR-10 dataset when $SNR_{\text{test}} \neq SNR_{\text{train}}$. Top row: CPP=1/6, bottom row: CPP=1/12.}   
         \label{fig:snr_mismatch_cifar}
\end{figure}

The performance curves when the training and testing SNR mismatch ranges from -5dB to 10dB are shown in Figure \ref{fig:snr_mismatch_cifar}. 
We also present visualization of the first 36 images of the CIFAR-10 test dataset in Figure \ref{fig:mismatch_visual_cifar}.
Since the codeword dimension of CIFAR-10 images is much smaller than that of Kodak images, the error recovery via ISEC in lower SNR environments is not significant (note there might be room to improve since we did not optimize the ISEC parameters $\alpha, \delta$, and $\eta$).
The suboptimal performance of one-shot decoding when the test SNR is higher than the training SNR is observed in Figure \ref{fig:snr_mismatch_cifar} and \ref{fig:cifar_higher_snr_blurry}. 
ISEC provides significant gains in such scenarios.

\begin{figure}
    \centering
    \includegraphics[width=1.0\linewidth]{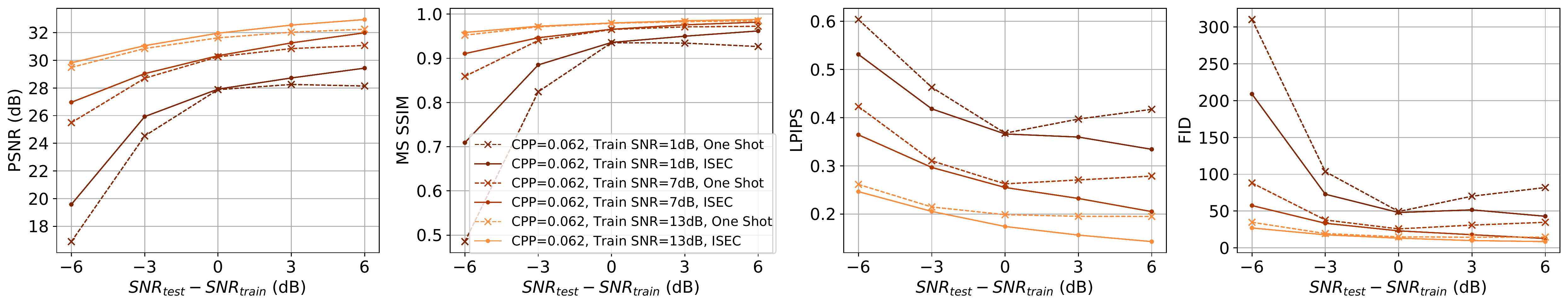}
    \caption{Reconstruction Performance of ISEC on Kodak dataset when $SNR_{\text{test}} \neq SNR_{\text{train}}$, CPP=1/16. The x-axis indicates the SNR difference between tested and trained environment. }
    \label{fig:snr_mismatch_nc6_kodak}
\end{figure}

Figure \ref{fig:snr_mismatch_nc6_kodak} shows the performance gain on Kodak dataset when $SNR_{\text{test}} \neq SNR_{\text{train}}$, CPP=1/16.
The same trend is observed as when CPP=1/6 (see Figure \ref{fig:snr_mismatch}).

\begin{figure}[]
    \centering
    \begin{subfigure}{0.32\linewidth}
        \centering
        \includegraphics[width=1.0\textwidth]{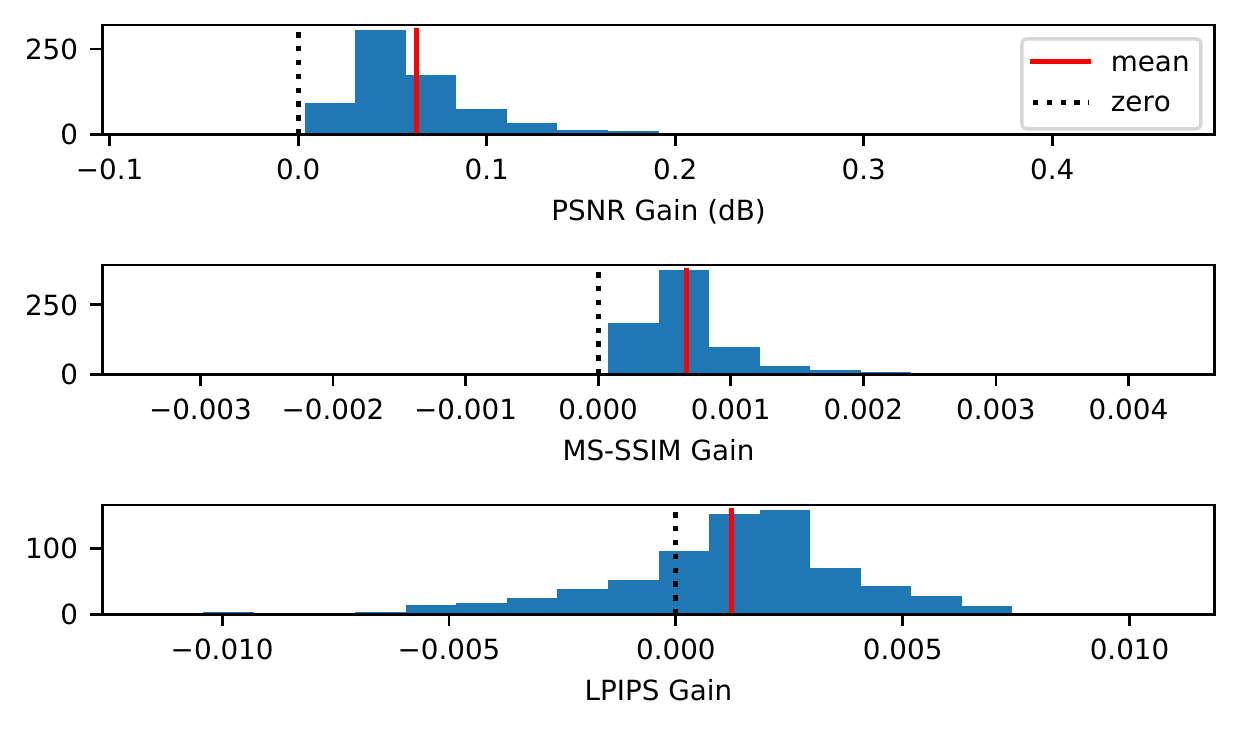}
        \caption{CPP=1/6, $SNR_{\text{test}}$=1dB}
    \end{subfigure}
    \begin{subfigure}{0.32\linewidth}
        \centering
        \includegraphics[width=1.0\textwidth]{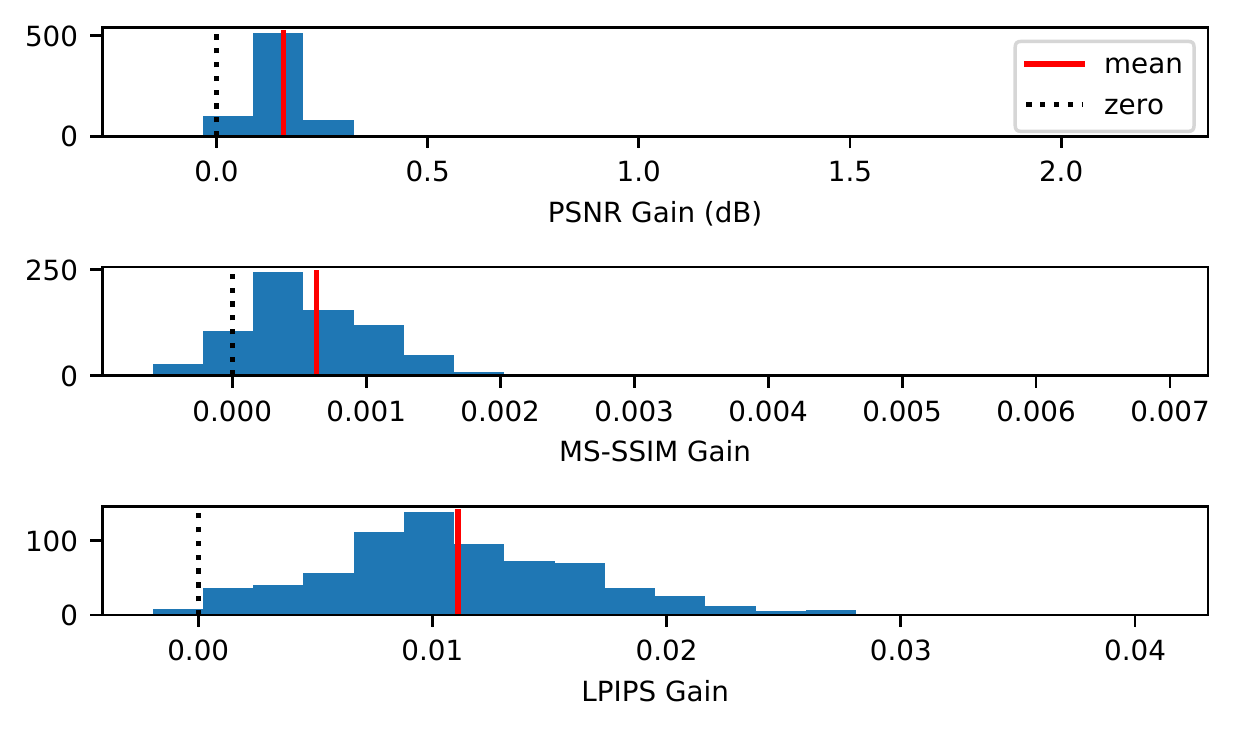}
        \caption{CPP=1/6, $SNR_{\text{test}}$=7dB}
    \end{subfigure}
    \begin{subfigure}{0.32\linewidth}
        \centering
        \includegraphics[width=1.0\textwidth]{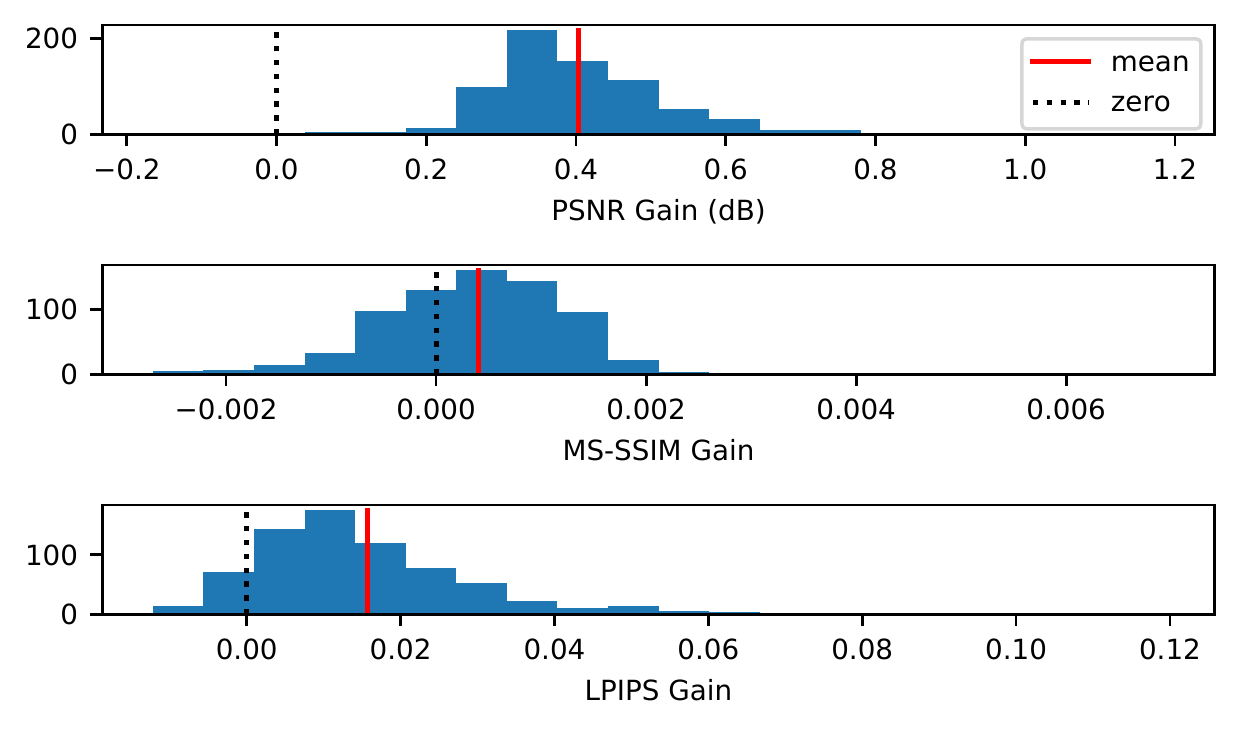}
        \caption{CPP=1/6, $SNR_{\text{test}}$=13dB}
    \end{subfigure}
    
    \begin{subfigure}{0.32\linewidth}
        \centering
        \includegraphics[width=1.0\textwidth]{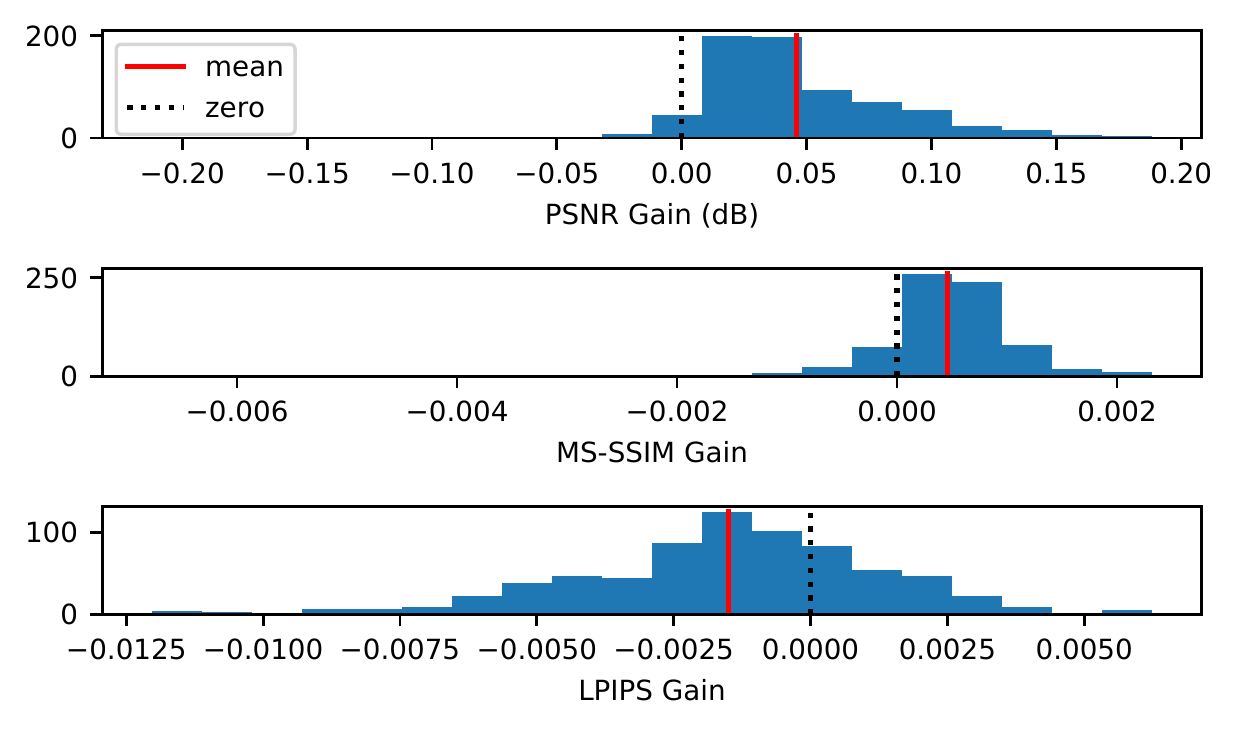}
        \caption{CPP=1/16, $SNR_{\text{test}}$=1dB}
    \end{subfigure}
    \begin{subfigure}{0.32\linewidth}
        \centering
        \includegraphics[width=1.0\textwidth]{Figures/Dist/kodak_dist_7_6.pdf}
        \caption{CPP=1/16, $SNR_{\text{test}}$=7dB}
    \end{subfigure}
    \begin{subfigure}{0.32\linewidth}
        \centering
        \includegraphics[width=1.0\textwidth]{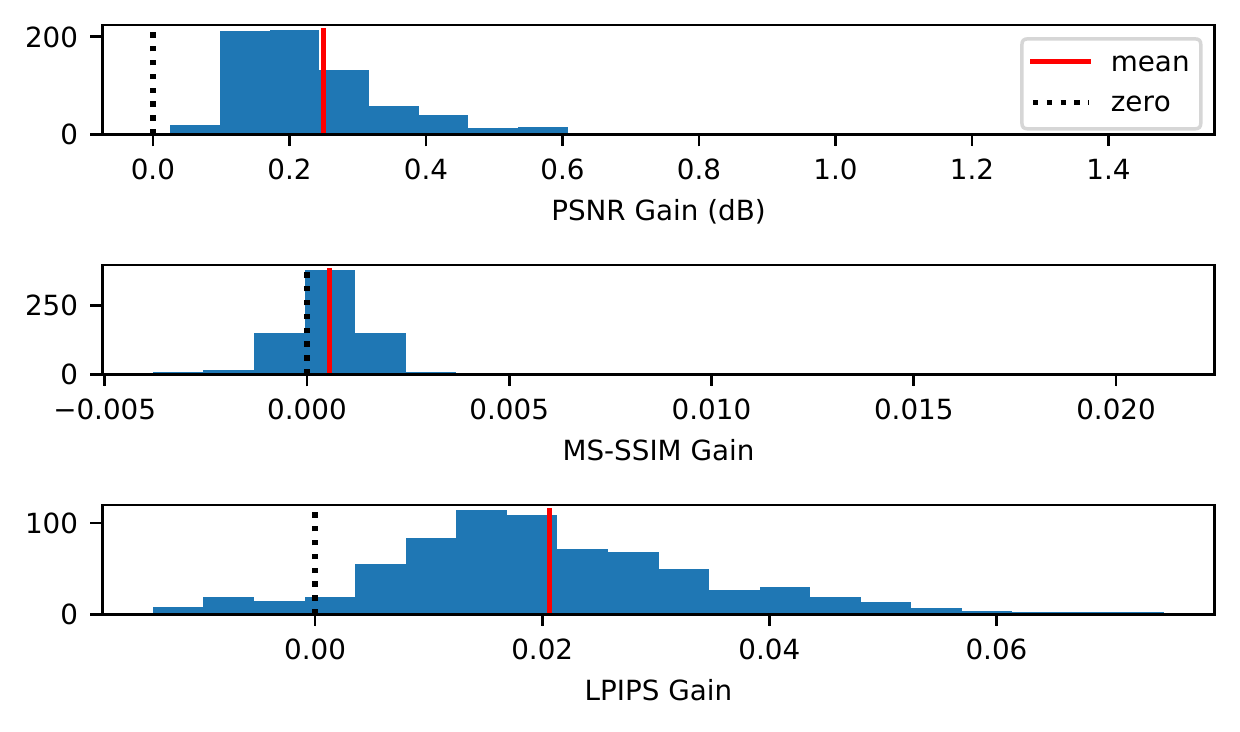}
        \caption{CPP=1/16, $SNR_{\text{test}}$=13dB}
    \end{subfigure}
    \caption{Histograms of ISEC gains for randomly cropped $256\times 256$ Kodak images using $SNR_{\text{train}}=7$dB model.}
    \label{fig:histogram_all}
\end{figure}

\subsection{Distributions of Gains from ISEC for Kodak images}
In this section, we show distributions of the gain from ISEC for three different test SNRs (1, 7, and 13 dB) at CPP=1/6 and 1/12 using the 7dB model.
It supplements the analysis shown in Figure \ref{fig:kodak_hist}.
Figure \ref{fig:histogram_all} shows histograms of PSNR, MS-SSIM, and LPIPS gains from ISEC for randomly cropped $256\times 256$ pixel Kodak image patches.
We noticed ISEC does not provide  LPIPS gains when CPP=1/16 and SNR=1dB (-6dB lower than the training SNR).
We presume the ISEC gain diminishes for this challenging scenario (compression rate is high at CPP=1/16 and SNR is much lower than expected) because the quality of Deep JSCC one-shot decoding degrades considerably so that the starting point of ISEC is far away from the local optima.
Nevertheless, the average gains of all other metrics are above zero without a significant number of outliers exhibiting negative gains.


\section{Additional Qualitative/Visual Comparisons}\label{sec:append_qualitative}

\begin{figure}
    \newcommand{\ratio}{0.30}
    \centering
    \begin{subfigure}{\ratio\linewidth}
        \centering\includegraphics[width=1.0\linewidth]{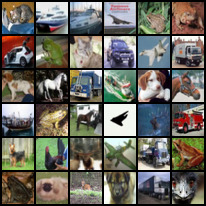}
        \caption{Target}
    \end{subfigure}
    \begin{subfigure}{\ratio\linewidth}
        \centering\includegraphics[width=1.0\linewidth]{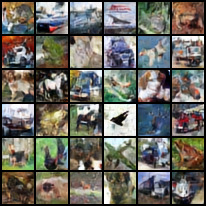}
        \caption{One-shot, $SNR_{\text{test}}=0$dB}
    \end{subfigure}   
    \begin{subfigure}{\ratio\linewidth}
        \centering\includegraphics[width=1.0\linewidth]{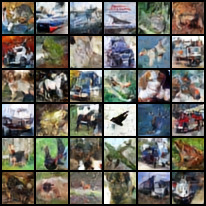}
        \caption{ISEC, $SNR_{\text{test}}=0$dB}
    \end{subfigure}    
    
    \begin{subfigure}{\ratio\linewidth}
        \centering\includegraphics[width=1.0\linewidth]{Figures/Images/targets_collection_st5_s0_nc16.jpg}
        \caption{Target}
    \end{subfigure}
    \begin{subfigure}{\ratio\linewidth}
        \centering\includegraphics[width=1.0\linewidth]{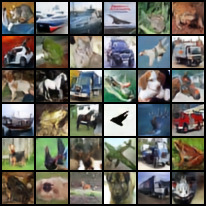}
        \caption{One-shot, $SNR_{\text{test}}=5$dB}
    \end{subfigure}   
    \begin{subfigure}{\ratio\linewidth}
        \centering\includegraphics[width=1.0\linewidth]{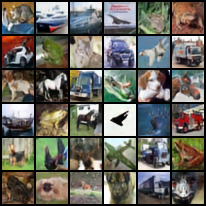}
        \caption{ISEC, $SNR_{\text{test}}=5$dB}
    \end{subfigure}    
    
    \begin{subfigure}{\ratio\linewidth}
        \centering\includegraphics[width=1.0\linewidth]{Figures/Images/targets_collection_st5_s0_nc16.jpg}
        \caption{Target}
    \end{subfigure}
    \begin{subfigure}{\ratio\linewidth}
        \centering\includegraphics[width=1.0\linewidth]{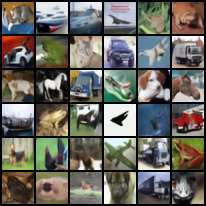}
        \caption{One-shot, $SNR_{\text{test}}=15$dB}
        \label{fig:cifar_higher_snr_blurry}
    \end{subfigure}   
    \begin{subfigure}{\ratio\linewidth}
        \centering\includegraphics[width=1.0\linewidth]{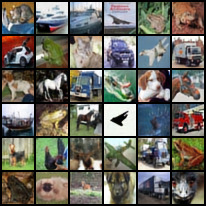}
        \caption{ISEC, $SNR_{\text{test}}=15$dB}
    \end{subfigure}    
    \caption{One-shot and ISEC decoding of CIFAR-10 test images using 5dB model on CPP=1/6.}
    \label{fig:mismatch_visual_cifar}
\end{figure}

\begin{figure*}[p]
\newcommand{\ratio}{0.3}
\centering
    \begin{subfigure}{\ratio\linewidth}
    \centering
    \includegraphics[width=\linewidth]{Figures/Images/st1.0_s-5.0_nc16_target0009.jpg}
    \includegraphics[width=\linewidth]{Figures/Images/st1.0_s-5.0_nc16_target0016.jpg}
    \caption{Target}
    \end{subfigure}
    \begin{subfigure}{\ratio\linewidth}
    \centering
    \includegraphics[width=\linewidth]{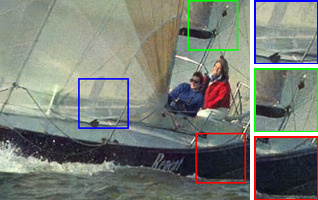}
    \includegraphics[width=\linewidth]{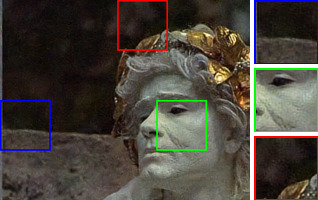}
    \caption{Before ISEC, $SNR_{\text{test}}=1$dB}
    \end{subfigure}
    \begin{subfigure}{\ratio\linewidth}
    \centering
    \includegraphics[width=\linewidth]{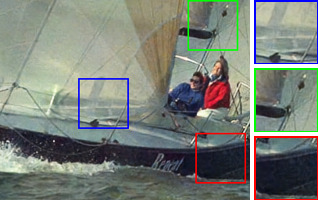}
    \includegraphics[width=\linewidth]{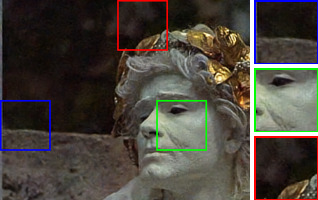}
    \caption{After ISEC, $SNR_{\text{test}}=1$dB}
    \end{subfigure}
    
    \begin{subfigure}{\ratio\linewidth}
    \centering
    \includegraphics[width=\linewidth]{Figures/Images/st1.0_s-5.0_nc16_target0009.jpg}
    \includegraphics[width=\linewidth]{Figures/Images/st1.0_s-5.0_nc16_target0016.jpg}
    \caption{Target}
    \end{subfigure}
    \begin{subfigure}{\ratio\linewidth}
    \centering
    \includegraphics[width=\linewidth]{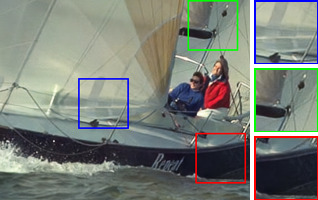}
    \includegraphics[width=\linewidth]{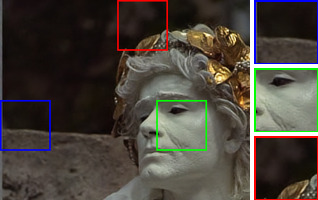}
    \caption{Before ISEC, $SNR_{\text{test}}=7$dB}
    \end{subfigure}
    \begin{subfigure}{\ratio\linewidth}
    \centering
    \includegraphics[width=\linewidth]{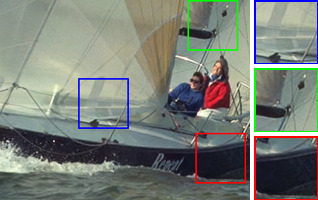}
    \includegraphics[width=\linewidth]{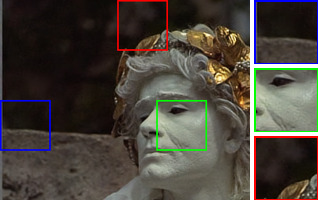}
    \caption{After ISEC, $SNR_{\text{test}}=7$dB}
    \end{subfigure}
    
    \begin{subfigure}{\ratio\linewidth}
    \centering
    \includegraphics[width=\linewidth]{Figures/Images/st1.0_s-5.0_nc16_target0009.jpg}
    \includegraphics[width=\linewidth]{Figures/Images/st1.0_s-5.0_nc16_target0016.jpg}
    \caption{Target}
    \end{subfigure}
    \begin{subfigure}{\ratio\linewidth}
    \centering
    \includegraphics[width=\linewidth]{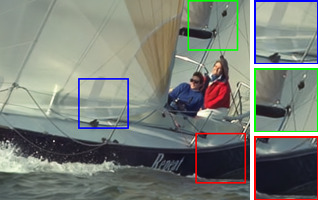}
    \includegraphics[width=\linewidth]{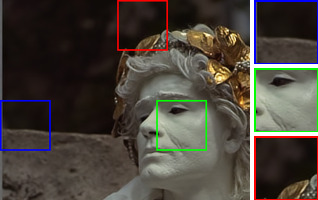}
    \caption{Before ISEC, $SNR_{\text{test}}=13$dB}
    \end{subfigure}
    \begin{subfigure}{\ratio\linewidth}
    \centering
    \includegraphics[width=\linewidth]{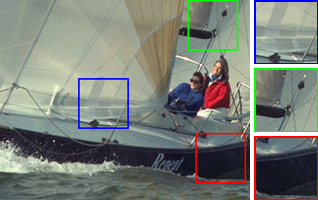}
    \includegraphics[width=\linewidth]{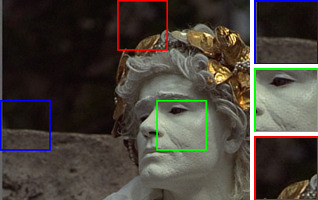}
    \caption{After ISEC, $SNR_{\text{test}}=13$dB}
    \end{subfigure}

    \caption{Before and After ISEC on Kodak images using $SNR_{\text{train}}$=7dB model on CPP=1/6.}
    \label{fig:mismatch_visual_cpp_1_6_7}
\end{figure*}

\begin{figure*}[p]
\newcommand{\ratio}{0.3}
\centering
   
    \begin{subfigure}{\ratio\linewidth}
    \centering
    \includegraphics[width=\linewidth]{Figures/Images/st1.0_s-5.0_nc16_target0009.jpg}
    \includegraphics[width=\linewidth]{Figures/Images/st1.0_s-5.0_nc16_target0016.jpg}
    \caption{Target}
    \end{subfigure}
    \begin{subfigure}{\ratio\linewidth}
    \centering
    \includegraphics[width=\linewidth]{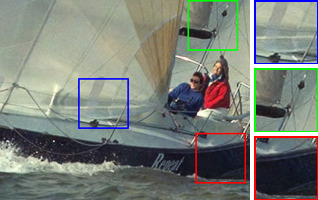}
    \includegraphics[width=\linewidth]{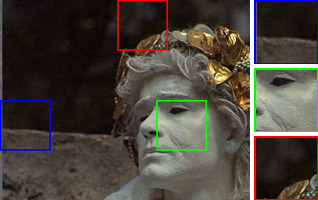}
    \caption{Before ISEC, $SNR_{\text{test}}=7$dB}
    \end{subfigure}
    \begin{subfigure}{\ratio\linewidth}
    \centering
    \includegraphics[width=\linewidth]{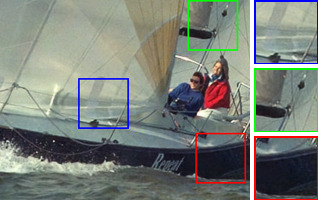}
    \includegraphics[width=\linewidth]{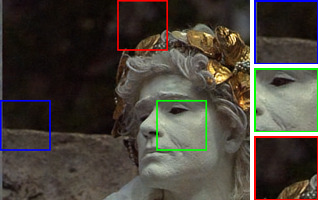}
    \caption{After ISEC, $SNR_{\text{test}}=7$dB}
    \end{subfigure}
    
    \begin{subfigure}{\ratio\linewidth}
    \centering
    \includegraphics[width=\linewidth]{Figures/Images/st1.0_s-5.0_nc16_target0009.jpg}
    \includegraphics[width=\linewidth]{Figures/Images/st1.0_s-5.0_nc16_target0016.jpg}
    \caption{Target}
    \end{subfigure}
    \begin{subfigure}{\ratio\linewidth}
    \centering
    \includegraphics[width=\linewidth]{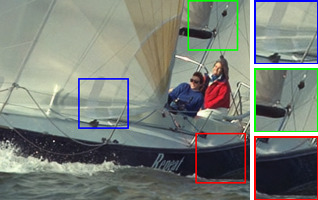}
    \includegraphics[width=\linewidth]{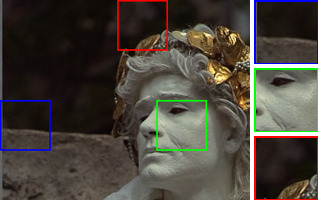}
    \caption{Before ISEC, $SNR_{\text{test}}=13$dB}
    \end{subfigure}
    \begin{subfigure}{\ratio\linewidth}
    \centering
    \includegraphics[width=\linewidth]{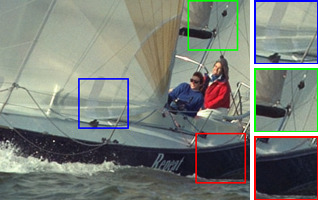}
    \includegraphics[width=\linewidth]{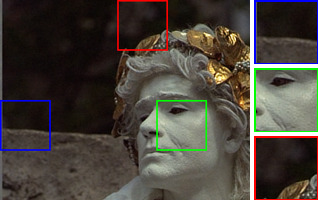}
    \caption{After ISEC, $SNR_{\text{test}}=13$dB}
    \end{subfigure}
     \begin{subfigure}{\ratio\linewidth}
    \centering
    \includegraphics[width=\linewidth]{Figures/Images/st1.0_s-5.0_nc16_target0009.jpg}
    \includegraphics[width=\linewidth]{Figures/Images/st1.0_s-5.0_nc16_target0016.jpg}
    \caption{Target}
    \end{subfigure}
    \begin{subfigure}{\ratio\linewidth}
    \centering
    \includegraphics[width=\linewidth]{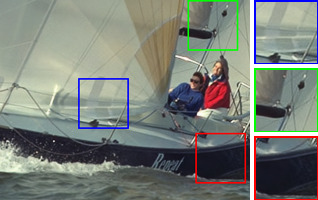}
    \includegraphics[width=\linewidth]{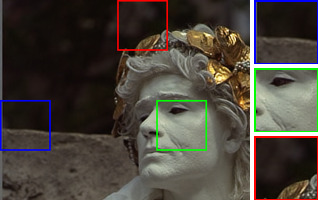}
    \caption{Before ISEC, $SNR_{\text{test}}=19$dB}
    \end{subfigure}
    \begin{subfigure}{\ratio\linewidth}
    \centering
    \includegraphics[width=\linewidth]{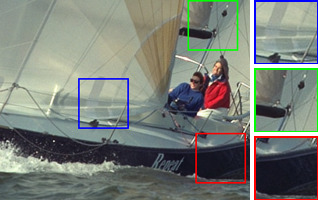}
    \includegraphics[width=\linewidth]{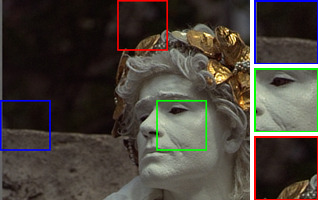}
    \caption{After ISEC, $SNR_{\text{test}}=1$dB}
    \end{subfigure}

    \caption{Before and After ISEC on Kodak images using $SNR_{\text{train}}$=13dB model on CPP=1/6.}
    \label{fig:mismatch_visual_cpp_1_6_13}
\end{figure*}

\begin{figure*}[p]
\newcommand{\ratio}{0.3}
\centering
    \begin{subfigure}{\ratio\linewidth}
    \centering
    \includegraphics[width=\linewidth]{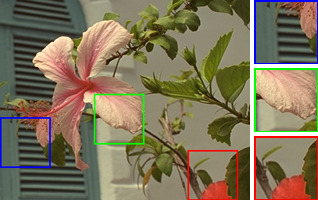}
    \includegraphics[width=\linewidth]{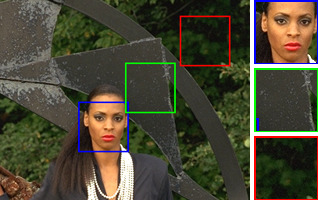}
    \caption{Target}
    \end{subfigure}
    \begin{subfigure}{\ratio\linewidth}
    \centering
    \includegraphics[width=\linewidth]{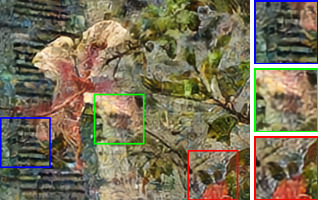}
    \includegraphics[width=\linewidth]{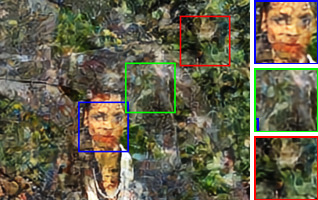}
    \caption{Before ISEC, $SNR_{\text{test}}=-5$dB}
    \end{subfigure}
    \begin{subfigure}{\ratio\linewidth}
    \centering
    \includegraphics[width=\linewidth]{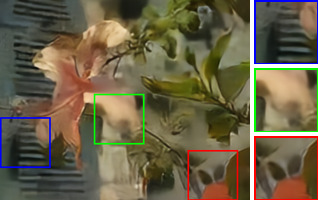}
    \includegraphics[width=\linewidth]{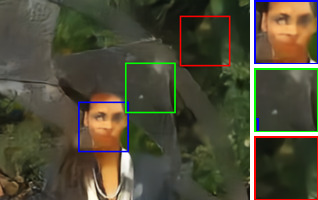}
    \caption{After ISEC, $SNR_{\text{test}}=-5$dB}
    \end{subfigure}
    
    \begin{subfigure}{\ratio\linewidth}
    \centering
    \includegraphics[width=\linewidth]{Figures/Images/st1.0_s-5.0_nc6_target0006.jpg}
    \includegraphics[width=\linewidth]{Figures/Images/st1.0_s-5.0_nc6_target0017.jpg}
    \caption{Target}
    \end{subfigure}
    \begin{subfigure}{\ratio\linewidth}
    \centering
    \includegraphics[width=\linewidth]{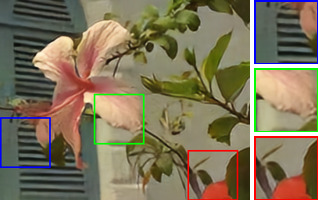}
    \includegraphics[width=\linewidth]{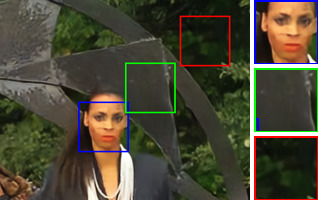}
    \caption{Before ISEC, $SNR_{\text{test}}=1$dB}
    \end{subfigure}
    \begin{subfigure}{\ratio\linewidth}
    \centering
    \includegraphics[width=\linewidth]{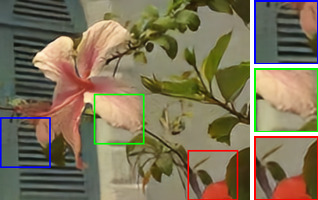}
    \includegraphics[width=\linewidth]{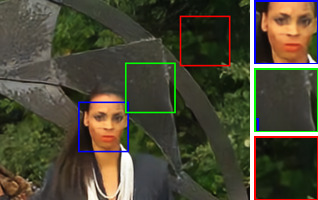}
    \caption{After ISEC, $SNR_{\text{test}}=1$dB}
    \end{subfigure}
    
    \begin{subfigure}{\ratio\linewidth}
    \centering
    \includegraphics[width=\linewidth]{Figures/Images/st1.0_s-5.0_nc6_target0006.jpg}
    \includegraphics[width=\linewidth]{Figures/Images/st1.0_s-5.0_nc6_target0017.jpg}
    \caption{Target}
    \end{subfigure}
    \begin{subfigure}{\ratio\linewidth}
    \centering
    \includegraphics[width=\linewidth]{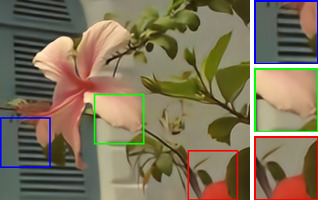}
    \includegraphics[width=\linewidth]{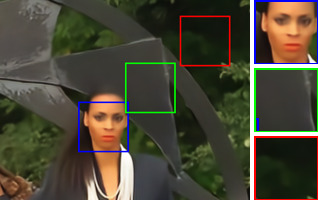}
    \caption{Before ISEC, $SNR_{\text{test}}=7$dB}
    \end{subfigure}
    \begin{subfigure}{\ratio\linewidth}
    \centering
    \includegraphics[width=\linewidth]{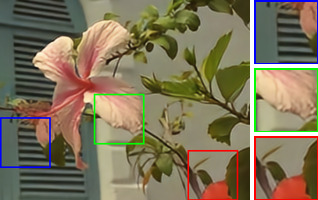}
    \includegraphics[width=\linewidth]{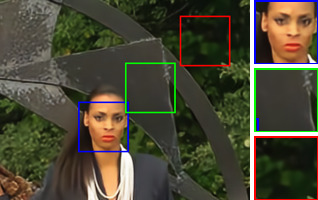}
    \caption{After ISEC, $SNR_{\text{test}}=7$dB}
    \end{subfigure}

    \caption{Before and After ISEC on Kodak images using $SNR_{\text{train}}$=1dB model on CPP=1/16.}
    \label{fig:mismatch_visual_cpp_1_16_1}
\end{figure*}

\begin{figure*}[p]
\newcommand{\ratio}{0.3}
\centering
    \begin{subfigure}{\ratio\linewidth}
    \centering
    \includegraphics[width=\linewidth]{Figures/Images/st1.0_s-5.0_nc6_target0006.jpg}
    \includegraphics[width=\linewidth]{Figures/Images/st1.0_s-5.0_nc6_target0017.jpg}
    \caption{Target}
    \end{subfigure}
    \begin{subfigure}{\ratio\linewidth}
    \centering
    \includegraphics[width=\linewidth]{Figures/Images/st7.0_s1.0_nc6_orig0006.jpg}
    \includegraphics[width=\linewidth]{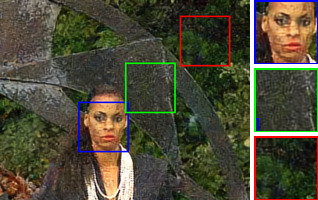}
    \caption{Before ISEC, $SNR_{\text{test}}=1$dB}
    \end{subfigure}
    \begin{subfigure}{\ratio\linewidth}
    \centering
    \includegraphics[width=\linewidth]{Figures/Images/st7.0_s1.0_nc6_updated0006.jpg}
    \includegraphics[width=\linewidth]{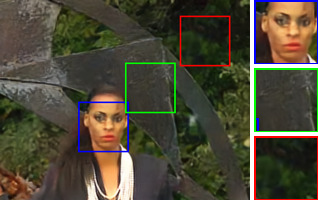}
    \caption{After ISEC, $SNR_{\text{test}}=1$dB}
    \end{subfigure}
    
    \begin{subfigure}{\ratio\linewidth}
    \centering
    \includegraphics[width=\linewidth]{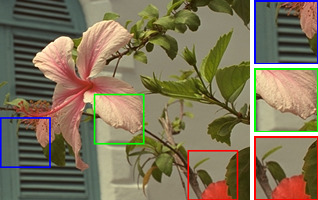}
    \includegraphics[width=\linewidth]{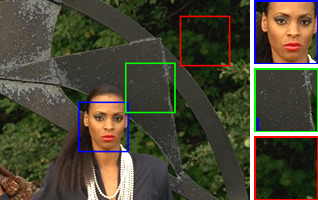}
    \caption{Target}
    \end{subfigure}
    \begin{subfigure}{\ratio\linewidth}
    \centering
    \includegraphics[width=\linewidth]{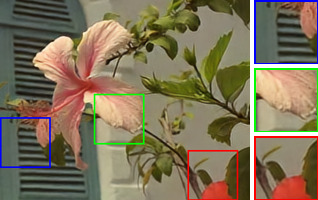}
    \includegraphics[width=\linewidth]{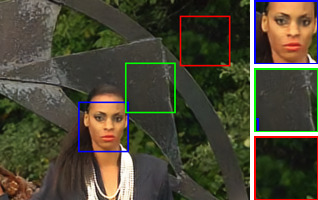}
    \caption{Before ISEC, $SNR_{\text{test}}=7$dB}
    \end{subfigure}
    \begin{subfigure}{\ratio\linewidth}
    \centering
    \includegraphics[width=\linewidth]{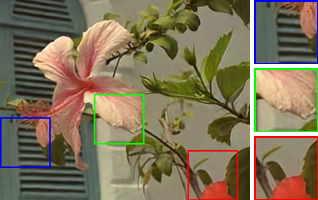}
    \includegraphics[width=\linewidth]{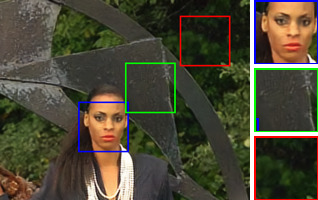}
    \caption{After ISEC, $SNR_{\text{test}}=7$dB}
    \end{subfigure}
    
    \begin{subfigure}{\ratio\linewidth}
    \centering
    \includegraphics[width=\linewidth]{Figures/Images/st1.0_s-5.0_nc6_target0006.jpg}
    \includegraphics[width=\linewidth]{Figures/Images/st1.0_s-5.0_nc6_target0017.jpg}
    \caption{Target}
    \end{subfigure}
    \begin{subfigure}{\ratio\linewidth}
    \centering
    \includegraphics[width=\linewidth]{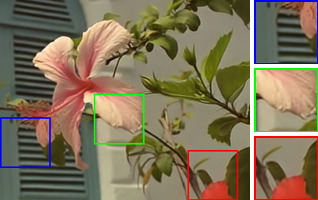}
    \includegraphics[width=\linewidth]{Figures/Images/st7.0_s13.0_nc6_orig0017.jpg}
    \caption{Before ISEC, $SNR_{\text{test}}=13$dB}
    \end{subfigure}
    \begin{subfigure}{\ratio\linewidth}
    \centering
    \includegraphics[width=\linewidth]{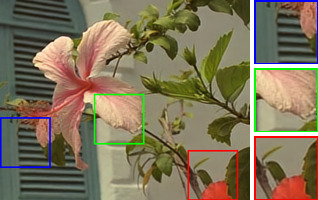}
    \includegraphics[width=\linewidth]{Figures/Images/st7.0_s13.0_nc6_updated0017.jpg}
    \caption{After ISEC, $SNR_{\text{test}}=13$dB}
    \end{subfigure}

    \caption{Before and After ISEC on Kodak images using $SNR_{\text{train}}$=7dB model on CPP=1/16.}
    \label{fig:mismatch_visual_cpp_1_16_7}
\end{figure*}

\begin{figure*}[p]
\newcommand{\ratio}{0.3}
\centering
   
    \begin{subfigure}{\ratio\linewidth}
    \centering
    \includegraphics[width=\linewidth]{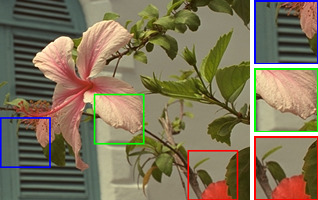}
    \includegraphics[width=\linewidth]{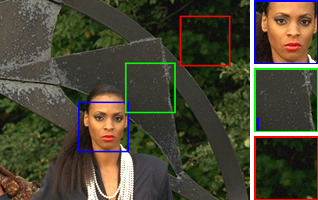}
    \caption{Target}
    \end{subfigure}
    \begin{subfigure}{\ratio\linewidth}
    \centering
    \includegraphics[width=\linewidth]{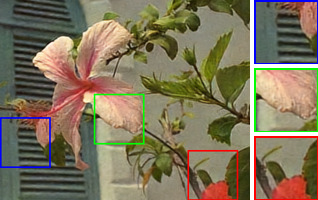}
    \includegraphics[width=\linewidth]{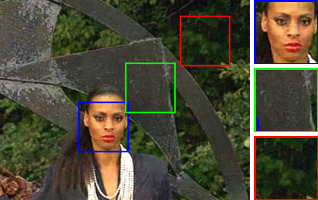}
    \caption{Before ISEC, $SNR_{\text{test}}=7$dB}
    \end{subfigure}
    \begin{subfigure}{\ratio\linewidth}
    \centering
    \includegraphics[width=\linewidth]{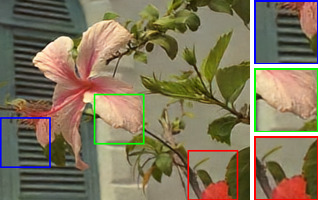}
    \includegraphics[width=\linewidth]{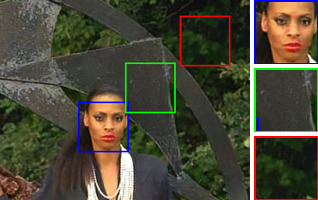}
    \caption{After ISEC, $SNR_{\text{test}}=7$dB}
    \end{subfigure}
    
    \begin{subfigure}{\ratio\linewidth}
    \centering
    \includegraphics[width=\linewidth]{Figures/Images/st1.0_s-5.0_nc6_target0006.jpg}
    \includegraphics[width=\linewidth]{Figures/Images/st1.0_s-5.0_nc6_target0017.jpg}
    \caption{Target}
    \end{subfigure}
    \begin{subfigure}{\ratio\linewidth}
    \centering
    \includegraphics[width=\linewidth]{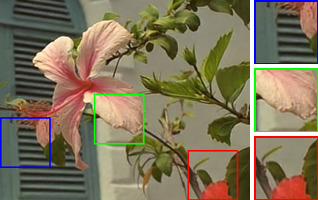}
    \includegraphics[width=\linewidth]{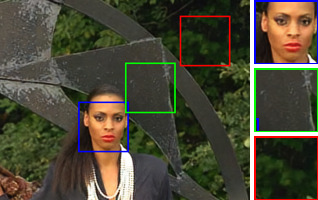}
    \caption{Before ISEC, $SNR_{\text{test}}=13$dB}
    \end{subfigure}
    \begin{subfigure}{\ratio\linewidth}
    \centering
    \includegraphics[width=\linewidth]{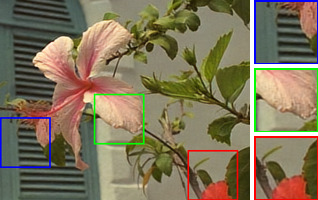}
    \includegraphics[width=\linewidth]{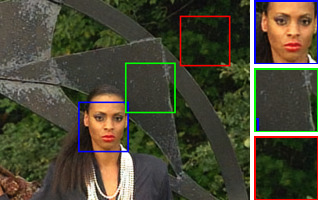}
    \caption{After ISEC, $SNR_{\text{test}}=13$dB}
    \end{subfigure}
     \begin{subfigure}{\ratio\linewidth}
    \centering
    \includegraphics[width=\linewidth]{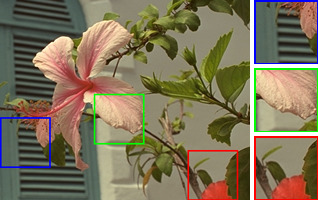}
    \includegraphics[width=\linewidth]{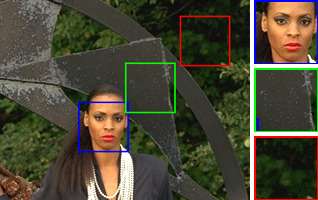}
    \caption{Target}
    \end{subfigure}
    \begin{subfigure}{\ratio\linewidth}
    \centering
    \includegraphics[width=\linewidth]{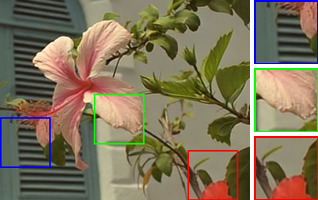}
    \includegraphics[width=\linewidth]{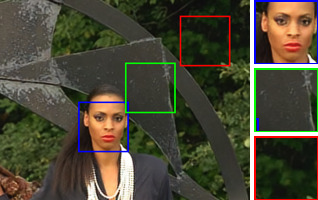}
    \caption{Before ISEC, $SNR_{\text{test}}=19$dB}
    \end{subfigure}
    \begin{subfigure}{\ratio\linewidth}
    \centering
    \includegraphics[width=\linewidth]{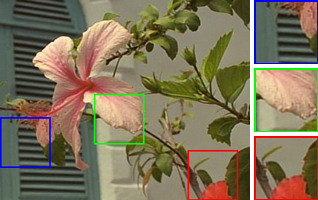}
    \includegraphics[width=\linewidth]{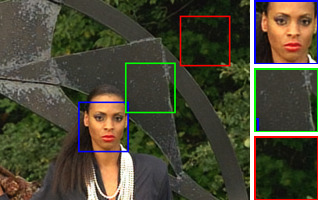}
    \caption{After ISEC, $SNR_{\text{test}}=1$dB}
    \end{subfigure}

    \caption{Before and After ISEC on Kodak images using $SNR_{\text{train}}$=13dB model on CPP=1/16.}
    \label{fig:mismatch_visual_cpp_1_16_13}
\end{figure*}

Figure \ref{fig:mismatch_visual_cifar} shows additional CIFAR-10 images decoded by one-shot decoding and ISEC based on the 5dB Deep JSCC model with CPP=1/6. Additional Kodak dataset images for one-shot decoding and ISEC results are shown in the following figures: Figure \ref{fig:mismatch_visual_cpp_1_6_1}, \ref{fig:mismatch_visual_cpp_1_6_7}, and \ref{fig:mismatch_visual_cpp_1_6_13} are at CPP=1/6 for 1, 7, and 13dB models, respectively.
Figure \ref{fig:mismatch_visual_cpp_1_16_1}, \ref{fig:mismatch_visual_cpp_1_16_7}, and \ref{fig:mismatch_visual_cpp_1_16_13} are outputs from 1, 7, and 13dB models, respectively, at CPP=1/16.


\begin{figure*}[p]
\newcommand{\ratio}{0.3}
\centering
    \begin{subfigure}{\ratio\linewidth}
    \centering
    \includegraphics[width=\linewidth]{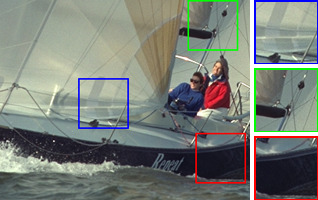}
    \includegraphics[width=\linewidth]{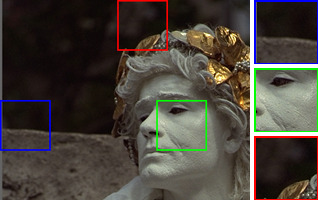}
    \caption{Target}
    \end{subfigure}
    \begin{subfigure}{\ratio\linewidth}
    \centering
    \includegraphics[width=\linewidth]{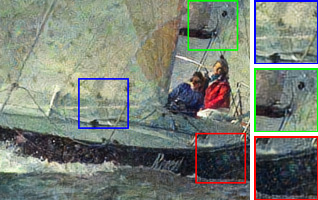}
    \includegraphics[width=\linewidth]{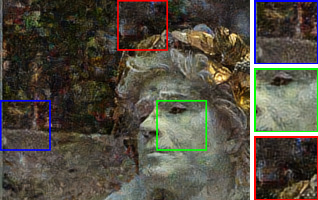}
    \caption{Before ISEC, $SNR_{\text{test}}=-5$dB}
    \end{subfigure}
    \begin{subfigure}{\ratio\linewidth}
    \centering
    \includegraphics[width=\linewidth]{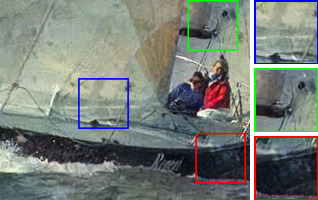}
    \includegraphics[width=\linewidth]{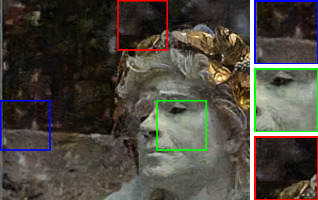}
    \caption{After ISEC, $SNR_{\text{test}}=-5$dB}
    \end{subfigure}
    
    \begin{subfigure}{\ratio\linewidth}
    \centering
    \includegraphics[width=\linewidth]{Figures/Images/st1.0_s-5.0_nc16_target0009.jpg}
    \includegraphics[width=\linewidth]{Figures/Images/st1.0_s-5.0_nc16_target0016.jpg}
    \caption{Target}
    \end{subfigure}
    \begin{subfigure}{\ratio\linewidth}
    \centering
    \includegraphics[width=\linewidth]{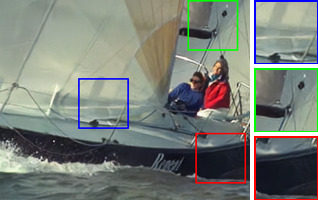}
    \includegraphics[width=\linewidth]{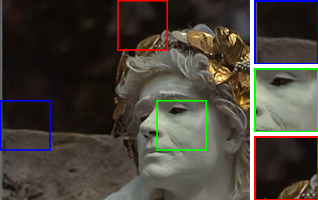}
    \caption{Before ISEC, $SNR_{\text{test}}=1$dB}
    \end{subfigure}
    \begin{subfigure}{\ratio\linewidth}
    \centering
    \includegraphics[width=\linewidth]{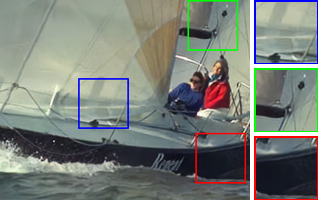}
    \includegraphics[width=\linewidth]{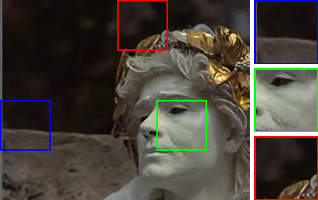}
    \caption{After ISEC, $SNR_{\text{test}}=1$dB}
    \end{subfigure}
    
    \begin{subfigure}{\ratio\linewidth}
    \centering
    \includegraphics[width=\linewidth]{Figures/Images/st1.0_s-5.0_nc16_target0009.jpg}
    \includegraphics[width=\linewidth]{Figures/Images/st1.0_s-5.0_nc16_target0016.jpg}
    \caption{Target}
    \end{subfigure}
    \begin{subfigure}{\ratio\linewidth}
    \centering
    \includegraphics[width=\linewidth]{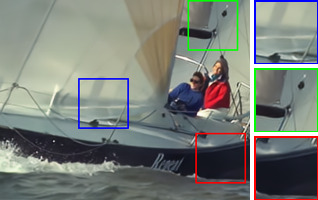}
    \includegraphics[width=\linewidth]{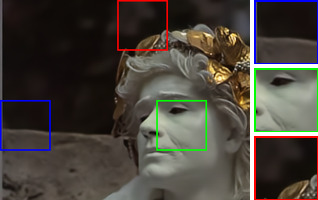}
    \caption{Before ISEC, $SNR_{\text{test}}=7$dB}
    \end{subfigure}
    \begin{subfigure}{\ratio\linewidth}
    \centering
    \includegraphics[width=\linewidth]{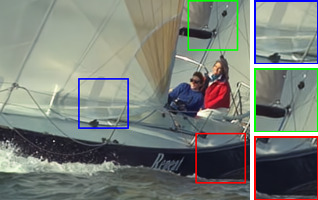}
    \includegraphics[width=\linewidth]{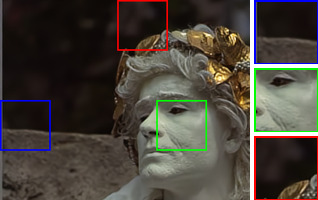}
    \caption{After ISEC, $SNR_{\text{test}}=7$dB}
    \end{subfigure}

    \caption{Before and After ISEC on Kodak images using $SNR_{\text{train}}$=1dB model on CPP=1/6.}
    \label{fig:mismatch_visual_cpp_1_6_1}
\end{figure*}

\end{document}